\pdfoutput=1

\documentclass[11pt]{article}

\usepackage[]{EACL2023}

\usepackage{times}
\usepackage{latexsym}

\usepackage[T1]{fontenc}

\usepackage[utf8]{inputenc}

\usepackage{microtype}

\usepackage{inconsolata}

\usepackage{array}
\usepackage{amsmath}
\usepackage{amssymb}
\usepackage{mathtools}
\usepackage{dsfont}
\usepackage{subcaption}
\usepackage{microtype}
\usepackage{booktabs}
\usepackage{colortbl}
\definecolor{myblue}{RGB}{31,119,180}
\definecolor{myred}{RGB}{214, 39, 40}
\definecolor{myorange}{RGB}{255, 127, 14}
\DeclarePairedDelimiter\abs{\lvert}{\rvert}
\DeclarePairedDelimiter\bigabs{\bigg\lvert}{\bigg\rvert}
\DeclareMathOperator{\freq}{freq}
\DeclareMathOperator{\acc}{acc}
\DeclareMathOperator{\conf}{conf}
\newcommand{\vc}{\text{victim counts }}
\newcommand{\ood}{\text{out-of-distribution }}
\newcommand{\EM}{\textit{Exact-Match}}
\newcommand{\fone}{$F_{1}$}

\usepackage{cleveref}
\usepackage[group-separator={,}]{siunitx}
\crefname{section}{\S}{\S\S}
\Crefname{section}{\S}{\S\S}
\crefname{table}{Tab.}{}
\crefname{figure}{Fig.}{}
\crefname{algorithm}{Algorithm}{}
\crefname{equation}{eq.}{}
\crefname{appendix}{App.}{}
\crefname{thm}{Theorem}{}
\crefname{prop}{Proposition}{}
\crefname{cor}{Corollary}{}
\crefname{observation}{Observation}{}
\crefname{assumption}{Assumption}{}
\crefformat{section}{\S#2#1#3}
\title{Extracting Victim Counts from Text}

\author{Mian Zhong ~\;~ ~\;~ Shehzaad Dhuliawala ~\;~ ~\;~ Niklas Stoehr \\
Institute for Machine Learning, ETH Z{\"u}rich \\
\href{mailto:mzhong@ethz.ch}{\small{\texttt{mzhong@ethz.ch}}} ~\;~
\href{mailto:shehzaad.dhuliawala@inf.ethz.ch}{\small{\texttt{shehzaad.dhuliawala@inf.ethz.ch}}} ~\;~
\href{mailto:niklas.stoehr@inf.ethz.ch}{\small{\texttt{niklas.stoehr@inf.ethz.ch}}} 
 }

\begin{document}
\maketitle
\begin{abstract}
Decision-makers in the humanitarian sector rely on timely and exact information during crisis events. Knowing how many civilians were injured during an earthquake is vital to allocate aids properly. Information about such \emph{victim counts} is often only available within full-text event descriptions from newspapers and other reports. Extracting numbers from text is challenging: numbers have different formats and may require numeric reasoning. This renders purely string matching-based approaches insufficient. As a consequence, fine-grained counts of injured, displaced, or abused victims beyond fatalities are often not extracted and remain unseen. We cast victim count extraction as a question answering (QA) task with a regression or classification objective. We compare regex, dependency parsing, semantic role labeling-based approaches, and advanced text-to-text models. Beyond model accuracy, we analyze extraction reliability and robustness which are key for this sensitive task. In particular, we discuss model calibration and investigate few-shot and out-of-distribution performance. Ultimately, we make a comprehensive recommendation on which model to select for different desiderata and data domains. Our work is among the first to apply numeracy-focused large language models in a real-world use case with a positive impact.\footnote{Code is available online at:\\ \url{https://github.com/mianzg/victim_counts}}
\end{abstract}

\section{Introduction}\label{sec:intro} 
Timely and accurate information during crisis events is crucial for rescue operations and the allocation of humanitarian aid~\cite{lepuschitz_seismographapi_2021}. However, crisis information is often scarce, subjective, or biased, which renders reported numbers in text extremely important~\cite{hellmeier_spotlight_2018, zavarella_mastering_2020, radford_automated_2021}.
For instance, the count of injured or missing people provides quantitative information about the catastrophic impact of an earthquake. 
In this work, we focus on human victims in crisis events, e.g., \ fatalities in floods, herein referred to as \emph{victim counts}. 
A reliable estimate of victim counts is helpful during crisis~\citep{Darcy_Hofmann_2003,Kreutzer_2020_improve}, and also post-crisis, benefiting research to diversify measures of crisis intensity. 
As of now, most intensity measures are either limited to event types~\cite{vincent_project_1979, goldstein_conflict-cooperation_1992}, fatality counts~\cite{kalyvas_logic_2006, chaudoin_beyond_2017} or both~\cite{stoehr_ordinal_2022}. 
More fine-grained measures such as injured, displaced, or abused victims are not captured in most popular databases and remain unmonitored~\citep{Krause_2013,Cruyff_van_Dijk_van_der_Heijden_2017,Cullen_Dawson_Price_Rowlands_2021}.

Many victim counts are reported in full-text form within event descriptions in news media. This makes their systematic collection and analysis technically complex. Manual extraction of victim counts from text is very labor-intensive and does not scale to big data collections~\citep{wad2016, navco3}. Computerized approaches such as the event coding software Tabari~\cite{schrodt_tabari_2009} and Petrarch2~\cite{Norris2017} focus on extracting actor and event types. They rely on lambda calculus and syntactic pattern matching, but disregard mentions of victim counts.
 
As we will show, parsing-based approaches perform decently well at extracting explicitly reported victim counts. They can identify the mention of the count ``$5$'' in ``5 people were injured''. However, they are often inadequate when the description \textit{implies} a correct count --- for example, from the description that ``one logger was shot but survived'', a human reader may infer that \emph{one} person is injured. Since neither a count nor the injury is mentioned explicitly, a parsing-based system may fall short. Another difficulty stems from the fact that the counts can be reported in many, different formats. A reported count may be digit-based or spelled out, define an exact quantity or a range as in ``dozens of people were injured''. As a consequence, formulating the task of victim count extraction is not an easy endeavor~(\cref{sec:task}). Most prior work assumes a setting where the count is explicitly mentioned in an event description~\cite{Dohling_Leser_2011,Imran_Elbassuoni_Castillo_Diaz_Meier_2013,Rudra_Ganguly_Goyal_Ghosh_2018,Camilleri_Azzopardi_Agius_2019}. Such settings can be tackled by sequence labeling models that select a relevant span from the given description. However, if the victim count does not appear verbatim, as in the above ``one logger'' example, models with some form of abstract reasoning capacity may be needed~\cite{roy_reasoning_2015}. Recently, large language models have shown promising results in answering number-focused questions with and without explicit mentions of relevant numbers~\cite{lewkowycz_solving_2022, nye_show_2022, wei_chain--thought_2022, lefebvre_rethinking_2022}. 

This paper is concerned with studying these different approaches~(\cref{sec:models}): as baselines, we compare regular expression, dependency parsing, and semantic role labeling. We consider the NT5~\cite{Yang2021NT5TT} model as a representative numeracy-enhanced pre-trained language model. We use the representation of this model in a generation, a classification, and a regression setting. We evaluate all models along three dimensions: accuracy~(\cref{sec:accuracy}), reliability~(\cref{sec:reliability}), and robustness~(\cref{sec:robustness}). We find that the fine-tuned language model outperforms the baseline models, especially when the victim count extraction requires reasoning. Reliability and robustness are particularly important in high-stake, human-centric tasks such as victim count extraction~\cite{zhang2020effect, kong-etal-2020-calibrated, russo_understanding_2022}. Model reliability indicates to which extent model behavior can be trusted within decision-making settings~\cite{leibig2017leveraging, Jiang_Araki_Ding_Neubig_2021}. One dimension of reliability is model calibration which indicates if a model's confidence is aligned well with it making correct predictions~\cite{guo_calibration_2017}. While calibration has been widely studied for classification, we add to the discussion of calibrated regression~\cite{pmlr-v97-song19a} and generation settings \cite{widmann2021calibration}. Finally, the dimension of robustness describes how stably a model performs. For instance, when the training set is limited or when the test data is out-of-distribution, a less robust model will forfeit more of its predictive performance. To shed light on this dimension, we conduct experiments in few-shot learning and \ood settings. 

We conclude with an application to showcase the extraction of fine-grained and highly specialized types of victim counts. Lastly, we discuss the benefits and drawbacks of the different approaches to assist practitioners in choosing the most suitable task formulation and model.

\section{Data}\label{sec:data}
We use publicly available datasets covering natural disasters and armed conflicts, namely: (1) \href{https://eventdata.parusanalytics.com/data.dir/atrocities.html}{World Atrocities Dataset (WAD)}~\citep{wad2016}, (2) \href{https://dataverse.harvard.edu/dataset.xhtml?persistentId=doi:10.7910/DVN/INNYEO}{Non-violent and Violent Campaigns and Outcomes 3.0~(NAVCO)}~\citep{navco3}, and (3) \href{https://knowledge4policy.ec.europa.eu/online-resource/europe-media-monitor-emm_en}{European Media Monitor (EMM)}~\citep{emm, emm2}.
For each dataset, we use the event text description and two types of victim counts: the death count and the injury count that we refer to as ``WAD death'' or ``WAD injury''. We pre-process the data by removing the samples with missing values~(NaN) in the \vc. For EMM, we only consider samples with a non-zero victim count since ``\num{0}'' is over-represented.

\section{Task Formulation}\label{sec:task}
In this section, we discuss some questions and challenges faced in formulating the task of extracting victim counts from event descriptions.
We justify some of the choices we make and describe why it is not possible to have a single formulation that fits all needs:

\paragraph{Is the victim count always present in the text?} 
Victim counts can be expressed in various ways in the text. 
When the count is expressed explicitly in the text, say ``5 people were injured'', a span extraction model can effectively extract the injury count $5$. 
However, in certain cases, a single explicit number might not be mentioned, and the victim count needs to be logically or algebraically inferred from the text.
Consider the description ``a 4-year-old girl and her mother were found dead''; a model would need to logically deduce that the victim count of death is $2$. 
To handle this, we not only look at span extraction models but also experiment with models that can understand the text at a deeper level and produce a victim count. 

\paragraph{Is the victim count always a single number?} 
Often, in the event description, the victim count is described as a range, such as ``at least 330 people died'', or in vague terms, like ``dozens were injured''. 
Additionally, even within a description, the victim counts for the same event can be varying, possibly because of recording the counts from different sources. 
This makes extracting a single exact count almost impossible.
In such cases, the best a model can do is to output a close estimate of the actual victim count.
Another solution would be to provide a range within which the count could lie.
For a humanitarian section deciding on the quantity of aid to be deployed, a range might suffice over a single exact count. 
To account for this, we also look at models that are trained to output a range by classifying the victim counts into a set of binned categories.

\begin{figure*}[htb!]
    \centering
    \resizebox{\textwidth}{!}{
    \includegraphics{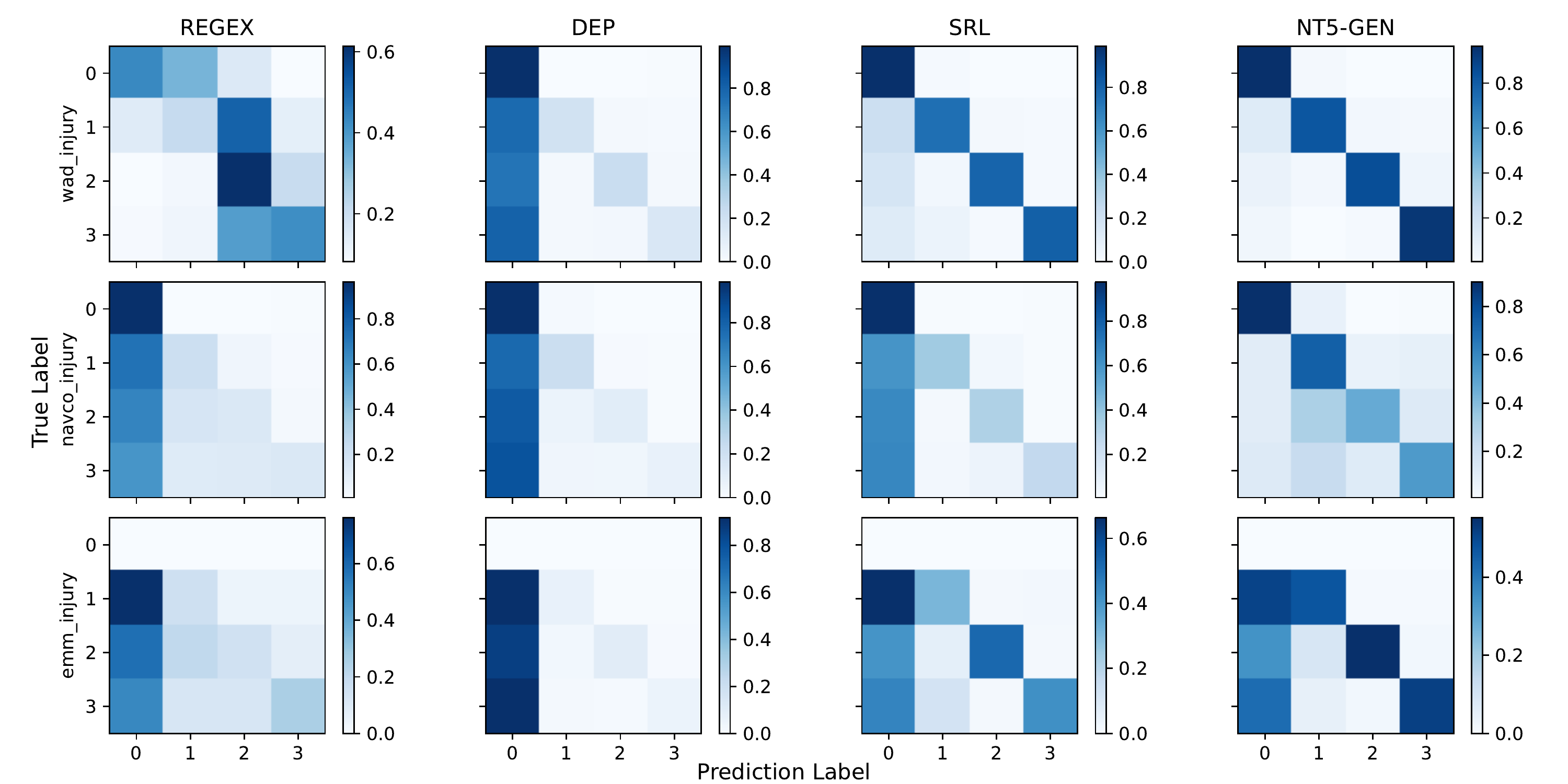}}
    \caption{Confusion matrices of the baselines and the fine-tuned NT5-Gen model (columns) of extracting injury counts from different data~(rows). We convert the true and prediction victim counts into 4 categories: for any count $y$, ``0'' is $y = 0$, ``1'' is $0< y \leq 3$, ``2'' is $3 < y \leq 10$ and  ``3'' is $y > 10$. Values are normalized over true counts. Baselines tend to have low precision on extracting injury counts (dark columns on ``0''). SRL and NT5-Gen have comparable accuracy and recall; however, NT5-Gen is slightly better in precision.}
    \label{fig:confusion_matrix_injury}
\end{figure*}

\section{Models}\label{sec:models}
In \cref{subsec:baselines}, we introduce baselines models that parse an event description and heuristically extract a victim count.
We then specify the model implementation for the different task formulations in \cref{subsec:model_implementation}. 

\subsection{Baseline Models}\label{subsec:baselines}
All baselines extract a victim count by locating the part of the text that could be relevant to victims and finding the nearby victim counts. 
The locating step requires a pre-defined list of words denoted as \emph{locating list}. For example, to extract death counts, this list would include terms like ``kill'' and ``die''.

\paragraph{Regex.} Regular expressions~(regex) is a rule-based method to extract counts by string pattern matching.
The patterns~(\cref{app:regex}) are built based on active or passive voice to extract a count closest to phrases in the locating list.

\paragraph{Dependency Parsing.} The dependency parsing model collects all possible numeric modifiers and their dependency relationships.
Since not every numeric modifier relates to victim counts, e.g., ``42-year-old'', we construct dependency rules with the locating list to decide if the number is the victim count.
For example, one rule is to check if the numeric modifier is for a subject phrase that would reject ``$42$'' in the example of ``42-year-old''.
If no numeric modifier is found~(e.g., ``a journalist was injured''), additional rules use the locating list to return ``$1$'' if the rule is satisfied and otherwise return ``$0$''. 

\paragraph{SRL.} Semantic role labeling~(SRL) recursively decomposes text input into pairs of predicates and their arguments.
We define a list of predicate verbs for death and injury count as the locating list.
Then, we iterate over the predicate-argument pairs, check if any predicate from the locating list occurs, and extract the count from its argument if possible.
If a predicate exists, the implementation returns the first number as the count if multiple are found and returns ``\num{1}'' if no verbatim number is found.
If no such predicate appears, the count is set to ``\num{0}''.

\subsection{Task Modeling}\label{subsec:model_implementation} 
We perform victim count extraction using three methods: generation, regression, and classification. 
As discussed above, each of these approaches caters to the different formulations of our task and can be beneficial in different scenarios.
Across these methods, we use the same underlying NT5 model. For clarity, we denote NT5-Gen, NT5-Reg, and NT5-Clf for the corresponding models. 
The NT5 model~\citep{Yang2021NT5TT} is a variant of the T5 model~\cite{raffel_exploring_2020} with further fine-tuning on numerical tasks.
We query the model in a similar fashion to previous works by giving the question and event description in the form: ``answer me:\texttt{[question]} context:\texttt{[passage]}''.
We discuss how we fine-tune this model for each of our specific methods below. 

\paragraph{Generation.} For generation, we fine-tune NT5 to decode the victim counts autoregressively. At inference, we use beam search to generate output. Generation does not guarantee to only generate numeral tokens; therefore, we follow \citet{cao2021autoregressive} to constrain the possible generation tokens in a prefix-conditioned way, such that only number digit tokens $0-9$ and \textsc{EOS} token are allowed at each decoding step.

\paragraph{Regression.}
For regression, we add two linear layers~(with ReLU activation) on the encoder representation to output the numerical victim count.
The model is trained to optimize the $\log$ mean-squared error between the true and predicted count.

\paragraph{Classification.} We model the task as a classification problem by binning the \vc into ordinal classes.
Similar to regression, the model has a classification head of a linear layer and a softmax layer on top of an encoder initialized with NT5 weights. 
Our experiments use a 3-class classification by converting the \vc into three categories: $[0, 3], (3, 10], (10, \infty)$.

\section{Accuracy of Counts Extraction}\label{sec:accuracy}
We begin by evaluating the efficacy of our proposed methods for victim count extraction. 
We examine the model accuracy by comparing baselines and the fine-tuned model with a generation objective~(\cref{subsec:compare_baseline}). We then show the results of using classification and regression formulations~(\cref{subsec:results-clf-regression}).

\begin{table}[htb!]
\fontsize{11}{11}\selectfont
\centering
\renewcommand{\arraystretch}{1.55} 
\setlength{\tabcolsep}{0.35em} 
\resizebox{\columnwidth}{!}{
\renewcommand{\arraystretch}{1.5}
\begin{tabular}{lrrrrrr}
\hline
 \multicolumn{1}{l}{} & \multicolumn{3}{c}{\textbf{\EM}} & \multicolumn{3}{c}{$\mathbf{F}_{1}$} \\
\cline{2-7}
 \multicolumn{1}{l}{} & WAD & NAVCO & EMM & WAD & NAVCO & EMM \\ 
\hline
  Regex & 0.117 & 0.264 & 0.064 & 0.202 & 0.318 & 0.124 \\ 
  Dep & 0.226 & 0.303 & 0.052 & 0.355 & 0.363 & 0.136 \\ 
  SRL & 0.741 & 0.430 & 0.313 & 0.779 & 0.484 & 0.361 \\ 
\hline
NT5-Gen & \textbf{0.813} & \textbf{0.501} & \textbf{0.443} & \textbf{0.846} & \textbf{0.544} & \textbf{0.492} \\
\hline
\end{tabular}}
\caption{\EM\space and \fone\space scores of the baseline models and the fine-tuned NT5-Gen on injury counts. The best results are \textbf{bolded}. The NT5-Gen model performs better than baselines across all datasets. \texttt{DEP} refers to the dependency parsing model and \texttt{SRL} refers to the semantic role labeling model.}
\label{tab:em_f1_injury}
\end{table}

\begin{table*}[htb!]
    \centering
    \renewcommand{\arraystretch}{1.1}
    \resizebox{\textwidth}{!}{
    \begin{tabular}{cp{7.8cm}rrr}
    \toprule
    Error Type & Context & Truth & SRL & NT5 \\
    \hline 
     \texttt{Diverse Expression} & \textit{Six passengers} in a taxi also \textit{\textcolor{myred}{had their throats cut}} & 6 & 0 & 6 \\
    \hline
    \texttt{Numerical Reasoning} &
          Herders shot and \textit{\textcolor{myred}{killed four people}} [...]. Herders then shot and \textit{\textcolor{myred}{killed a farmer}} at Jokhana [...]  & 5 & 4 & 5   \\
    \hline
    \texttt{Number Ambiguity} & \textit{\textcolor{myred}{Unidentified gunmen}} clash with army	& 1 & 0 & 1\\
    \hline
    \texttt{Number Spelling} & \textit{\textcolor{myred}{.}}Twenty\emph{\textcolor{myred}{-}}three people were killed [...] & 23 & 1 & 23 \\
    \end{tabular}}
    \caption{Error examples of SRL that the NT5-Gen model is correct on extracting death counts. \texttt{Diverse Expression} refers to the string patterns not captured by pre-defined rules. \texttt{Numerical Reasoning} shows that the correct count has to be achieved by some mathematical operation over the text. \texttt{Number Ambiguity} indicates that a verbatim number is not written but an estimate may be made (with domain expertise). \texttt{Number Spelling} refers to problems with number / text format that are typos or the tokenizer parses wrongly (e.g., ``twenty-three''$\rightarrow$ ``twenty'').}
    \label{tab:error_analysis}
\end{table*}

\subsection{Comparing Baselines with NT5-Gen}\label{subsec:compare_baseline}
We compare the accuracy performance of the baseline models and the fine-tuned NT5-Gen model.
\cref{tab:em_f1_injury} shows the results of extracting the injury counts using \EM\space and \fone scores commonly used in related tasks~\cite{Yang2021NT5TT,dua_drop_2019}. 
We measure \fone\space score on digitized tokens (i.e., ``$34$'' $\rightarrow$ [``$3$'', ``$4$'']). 
The fine-tuned NT5-Gen model has an accuracy boost up by $7$-$13\%$ in \EM\space and by $6$-$13\%$ in \fone\space score than the strongest baseline model SRL. 
The performance of regex and dependency parsing varies heavily across different data, which implies that the regex pattern or dependency relationship may be less helpful in finding the victim counts. 

Moreover, we convert the victim counts into four bins, where the bins are selected to have a balanced number of samples in each bin.
As an illustration, \cref{fig:confusion_matrix_injury} shows the confusion matrices on the transformed injury counts.
For both victim types, baseline models have a low precision to falsely return ``\num{0}'' too often. 
Compared with baselines, the NT5-Gen model improves to extract victim counts whose numeric values are large (e.g., $y>10$).

\paragraph{Qualitative Analysis.}
We qualitatively examine error samples of the SRL model that the NT5-Gen model extracts correctly.
We randomly select 20 error samples for each test set to evaluate and summarize 4 types of errors with examples in \cref{tab:error_analysis}.
Out of all errors\footnote{There are a few samples where the ground truth might be erroneous. As the event-coding requires more domain expertise within the corresponding social science discipline, we leave the discussion out of this work.}, $39.2\%$ belong to diverse linguistic expressions on depicting victims, $38.3\%$ contain number ambiguity, $8.3\%$ need numerical reasoning, and $5.8\%$ have spelling issues (for the tokenizer). 
The NT5-Gen model performs better when the count needs numerical reasoning.
Even if the reasoning is not needed, SRL may fail when the linguistic expression to depict victims (e.g., ``have throats cut'') is out of the pre-defined locating list~(e.g., [``die'', ``kill'', ``slay'']).
These error types are difficult for baseline models to be improved since the patterns cannot be defined beforehand.

\subsection{Results on Classification and Regression}\label{subsec:results-clf-regression}
\begin{figure}[htb!]
    \centering
    \resizebox{\columnwidth}{!}{
    \includegraphics{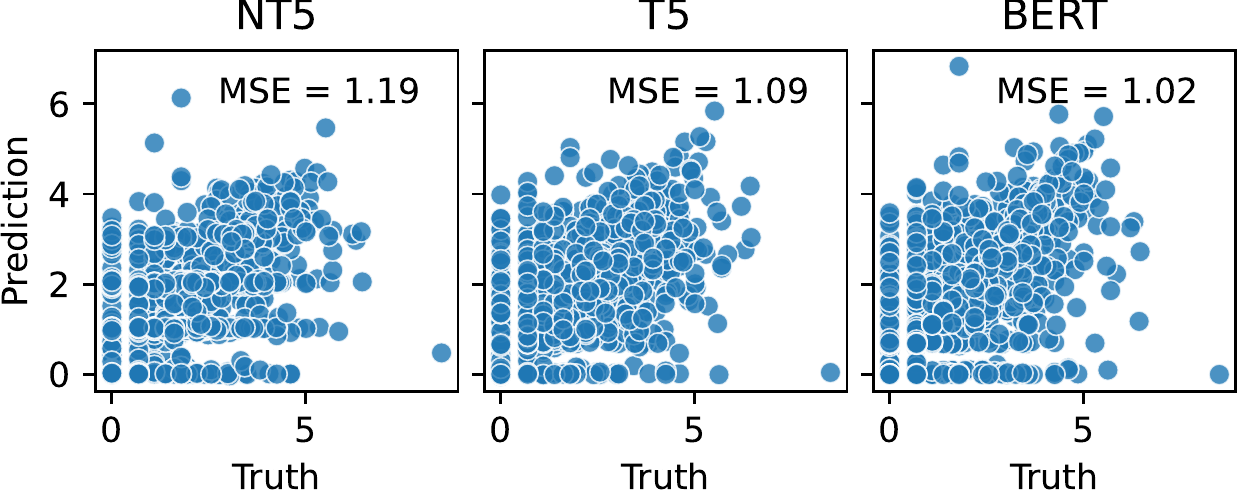}}
    \caption{Scatter plots of the fine-tuned NT5-Reg model initialized with different pre-trained weights (\textsc{nt5}, \textsc{t5-small}, and \textsc{bert-base-uncased}). The models are trained on log-transformed victim counts.}
    \label{fig:regression_navco_injury}
\end{figure}

We examine the accuracy of the classification and regression formulations by comparing NT5-Clf and NT5-Reg with different initialization weights. 
To compare, we use \textsc{t5-small} and \textsc{bert-base-uncased} pre-trained weights for the encoder.
\cref{tab:classification_accuracy} shows the classification results on NAVCO injury data.
Fine-tuning \textsc{t5-small} and \textsc{nt5} reaches comparable performance; precision and recall scores are similar, but precision is slightly higher. 
The scatter plots~(\cref{fig:regression_navco_injury}) show the results of regression using different pre-trained weights with the mean squared error~(MSE).
For a ($\log$-transformed) victim count larger than $5$, using the regression objective seems more conservative in giving small-valued predictions. 
The numeracy-rich NT5 weights do not particularly improve accuracy for a classification or regression objective, and employing some standard pre-trained weights might be sufficient.

\begin{table}[htb!]
\fontsize{11}{11}\selectfont
\centering
\renewcommand{\arraystretch}{1.} 
\setlength{\tabcolsep}{0.35em} 
\resizebox{\columnwidth}{!}{
\begin{tabular}{lrrrr}
\toprule
  & Accuracy & $\mathbf{F}_{1}$ & Precision & Recall\\
  \midrule
  \texttt{NT5} & 0.65 & 0.60 & 0.62 & 0.59\\
  \texttt{T5} & 0.65 & 0.60 & 0.61 & 0.59\\
  \texttt{BERT} & 0.52 & 0.23 & 0.17 & 0.33\\
\bottomrule
\end{tabular}}
\caption{Classification results on NAVCO injury data with the NT5-Clf model initialized by different pre-trained weights: \textsc{nt5}, \textsc{t5-small}, and \textsc{bert-base-uncased}. \fone\space, precision and recall scores are macro.}
\label{tab:classification_accuracy}
\end{table}

\section{Evaluating Reliability}\label{sec:reliability}
Another important dimension is reliability which we evaluate through the lens of calibration~(\cref{subsec:cal_metrics}). 
As we approach the task with multiple formulations, calibration analysis is especially needed to understand whether a model is calibrated~(\cref{subsec:cal_error}), and how post-hoc calibration techniques may adjust models to be better calibrated~(\cref{subsec:posthoc_cal}).

\subsection{Preliminaries: Calibration Metrics}{\label{subsec:cal_metrics}} 
A well-calibrated model ensures that the confidence of the output is well aligned with the chance of the output being accurate.
This is a desirable property for our task --- consider a model extracts ``0'' when the text depicts an injured person. 
A calibrated model would assign very low confidence to the extracted count, which may avoid error propagation to downstream decisions, e.g., medical resource dispatch. 
We here introduce the expected calibration error~($\mathrm{ECE}$)~\citep{Naeini2015ObtainingWC}, a standard metric used for classification and is extended for generation decoding~\cite{widmann2021calibration}. 
For regression, we apply quantile calibration error~\citep{pmlr-v80-kuleshov18a}.

Given $n$ samples, we create $M$ equal-width bins over the interval $[0,1]$.
ECE takes a weighted average on the differences between the classification accuracy and the mean confidence within each $B_m$,
\[\mathrm{ECE} = \sum_{m=1}^{M}\frac{\abs{B_m}}{n} \bigabs{\acc(B_m) - \conf(B_m)}. \] 
The quantile calibration error averages the differences between the empirical frequency $\freq(B_m)$ and the upper bound of $B_m$ (i.e., $\sup(B_m)$), where $\freq(B_m)$ is the fraction of $n$ samples whose quantiles lower or equal to $\sup(B_m)$,
\[\mathrm{RegCE} = \frac{1}{M}\sum_{m=1}^{M}\bigabs{\freq(B_m) - \sup(B_m)}.\] 

The calibration error of generation decoding takes the best $b$ beam search answers, and applies softmax on their scores to represent the confidence.
The ECE is then calculated on the best beam search answer similar to classification. 

\subsection{Calibration Error on Different Models}\label{subsec:cal_error}
\begin{table}[htb!]
\fontsize{9}{9}\selectfont
\centering
\renewcommand{\arraystretch}{1.25} 
\setlength{\tabcolsep}{0.35em} 
\resizebox{\columnwidth}{!}{
\begin{tabular}{clrrrr}
\toprule
  &  & \multicolumn{2}{c}{Death} & \multicolumn{2}{c}{Injury} \\
\midrule
  Data & Model & Orig & Calib. & Orig & Calib.  \\
\midrule
 NAVCO & Clf & 0.222 & 0.044 & 0.332 & 0.060\\ 
 & Reg & 0.220 & 0.097 & 0.141 & 0.057\\
 & Gen & 0.054 & 0.040 & 0.092 & 0.092\\ 
\toprule
 WAD & Clf & 0.192 & 0.055 & 0.228 & 0.088\\ 
 & Reg & 0.272 & 0.107 & 0.167 & 0.294 \\
 & Gen & 0.218 & 0.221 & 0.096 & 0.042\\
\toprule
 EMM & Clf & 0.277 & 0.098 & 0.314 & 0.055\\ 
 & Reg & 0.201 & 0.189 & 0.368 & 0.188 \\
 & Gen & 0.087 & 0.092 & 0.328 & 0.122\\
 \bottomrule
\end{tabular}}
\caption{
Calibration errors of fine-tuned NT5-Clf, NT5-Reg, and NT5-Gen models before (Orig.) and after (Calib.) applying post-hoc calibration. Post-hoc calibration effectively reduces the errors.
}
\label{tab:calibration_error}
\end{table}
We show in \cref{tab:calibration_error} the calibration errors measured on the fine-tuned NT5-Clf, NT5-Reg, and NT5-Gen with different data. 
Surprisingly, the NT5-Gen model is well-calibrated on most datasets, except for EMM injury: the lowest calibration error is $0.05$ on NAVCO death, and the errors on other data range between $0.08$ and $0.33$.
Classification models tend to have large calibration errors~($>0.19$). 
In particular, the error is larger than $0.3$ on NAVCO and EMM data to classify injury counts.
Regression is also prone to large calibration errors ($>0.15$).

\begin{figure}[htb!]
    \centering
    \resizebox{\columnwidth}{!}{
    \includegraphics{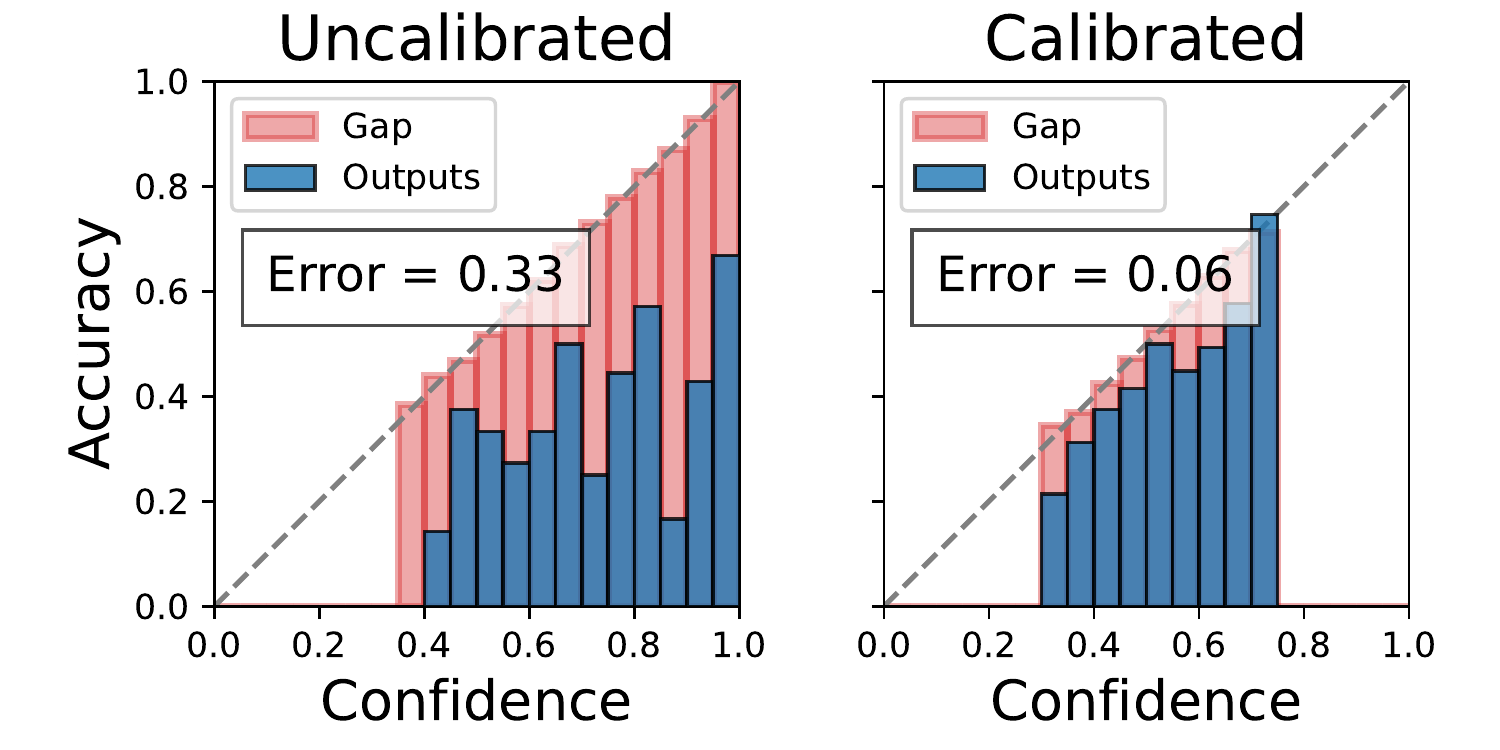}}
    \caption{Reliability diagrams compare the calibration error before~(left) and after~(right) post-hoc calibration of the fine-tuned NT5-Clf model using the NAVCO injury data. This model is prone to large calibration errors (red gaps) in many bins. This is especially true for bins with high model confidence ($>0.8$).
    }
    \label{fig:rd_clf_navco_injury}
\end{figure}

Another helpful tool is the reliability diagrams which visualize the calibration errors at different confidence bins.
As an illustration, \cref{fig:rd_clf_navco_injury} shows the diagram of the NT5-Clf model fine-tuned on NAVCO injury data, and the diagonal line indicates the perfect calibration.
This model is over-confident, and we can observe large gaps when the model confidence is larger than $0.8$.

\begin{figure*}[t!]
    \centering
    \resizebox{\textwidth}{!}{
    \includegraphics{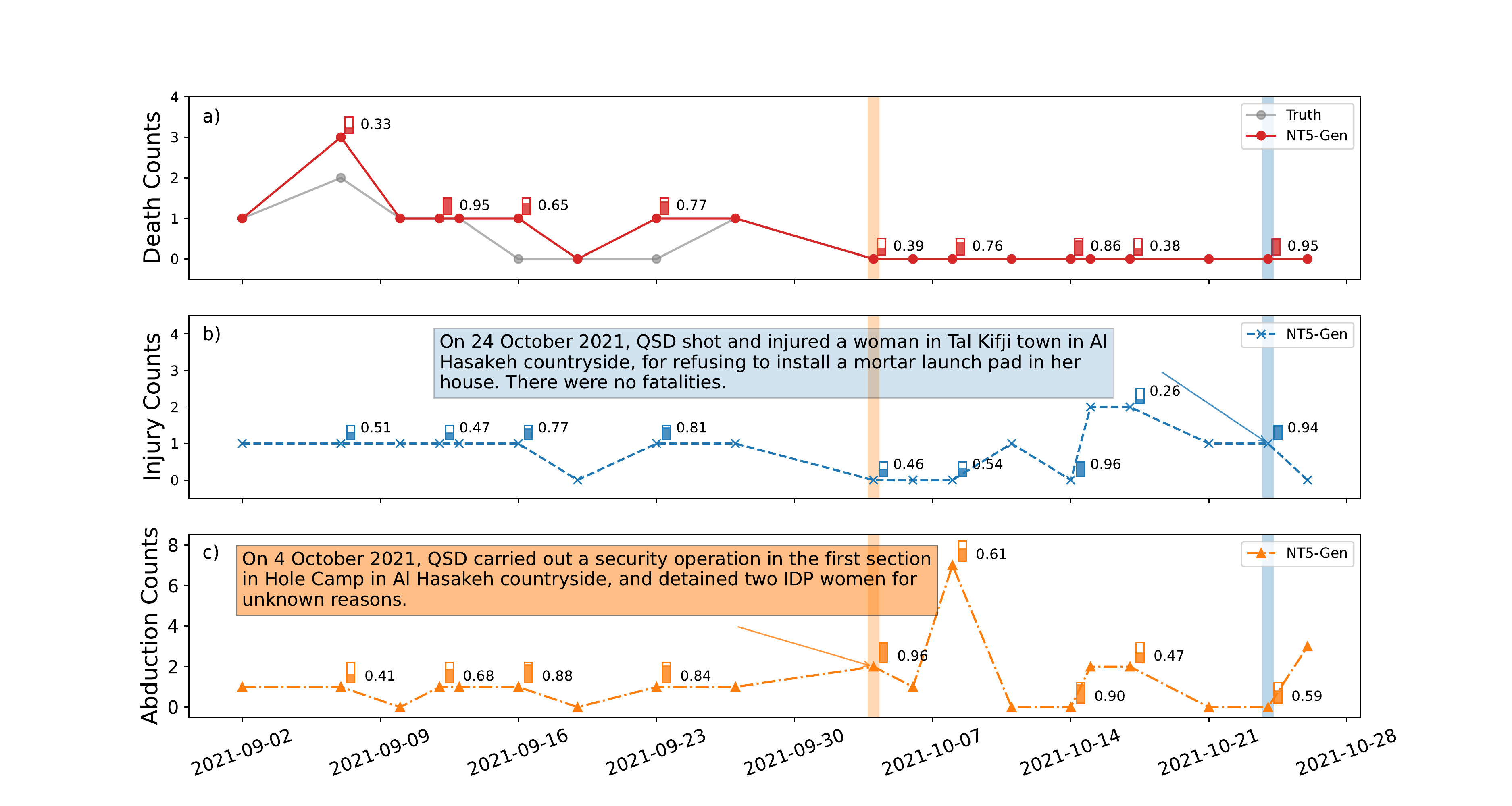}
    }
    \caption{Timeline of victim counts in Syria data from Sept to Nov 2021 as given in the ACLED dataset. We use the NT5-Gen model that is fine-tuned on NAVCO data. Our model can be tested on the extraction of \textcolor{myred}{fatality counts} which is the only victim count featured in ACLED (Fig. (a)). Beyond fatality counts, it can extract more fine-grained victim types such as (b) \textcolor{myblue}{injury} and (c) \textcolor{myorange}{abduction} counts. Confidence scores are shown for some of the predictions.}
    \label{fig:app}
\end{figure*}

\subsection{Post-hoc Calibration}\label{subsec:posthoc_cal}
Since the models can be over-confident based on the above analysis, we see the necessity to calibrate models for victim count extraction. 
We use temperature scaling for classification and generation decoding, and isotonic regression for regression.
The post-hoc calibrators use development data to minimize negative log-likelihood and are then applied to test sets to measure calibration errors.
As a comparison, \cref{fig:rd_clf_navco_injury}~(right) shows the calibrated results of the fine-tuned NT5-Clf model on NAVCO injury data. 
The calibration error~(i.e., ECE) reduces from $0.33$ to $0.06$.
The errors of other calibrated models can be found in \cref{tab:calibration_error}. 
In general, when the models have rather a large calibration error~(e.g., $>0.3$), post-hoc calibration is more helpful and adjusts the models to a better-calibrated level. 

\section{Evaluating Robustness}\label{sec:robustness}
Typically, conflict or disaster data is noisy and limited. This is making it challenging to train models on a large-scale, high-quality training set. For this reason, we need robust models that excel in few-shot and \ood settings.

\paragraph{Reduced Training Size.}
We fine-tune the NT5-Gen, NT5-Reg, and NT5-Clf models on different-size portions of the training set. Specifically, we use $100\%$, $50\%$, $10\%$,$ 5\%$, $0.5\%$ and $0\%$ of the training data and as further discussed in \cref{subsec:app_fewshot}.
As expected, we find that the accuracy of all models drops when using a smaller training set. 
The NT5-Gen model reveals to be the most robust in keeping the \EM\space metric above $0.6$ when being fine-tuned on only $5\%$ of the training data. 
The calibration error of the fine-tuned NT5-Clf model increases when the training size is reduced, while the fine-tuned NT5-Reg and NT5-Gen models do not follow this trend. 
In the zero-shot setting, the NT5-Reg and NT5-Gen models reach their largest calibration error. In contrast, the NT5-Clf model reaches its smallest calibration error in the zero-shot setting.

\paragraph{Out-of-distribution~(OOD) Setting.}
We set up synthetic tasks in which a fine-tuned model is confronted with an out-of-distribution setting at test time. For example, we fine-tune a model on WAD death and then repurpose it to classify WAD injury. Then, we evaluate the drop in performance of this ``out-of-distribution'' model compared to an ``in-distribution'' model, that has been trained on WAD injury labels directly. We conduct this comparison on different datasets and models.

In \cref{subsec:app_ood}, we evaluate the NT5-Clf model in a classification formulation and report accuracy. As expected, we find that accuracy decreases in every setting with performance drops between $0.001\%$ and $0.3\%$. In \cref{fig:reg_ood}, we evaluate the NT5-Reg model in a regression setting measured in MSE. We find that the performance decreases in the \ood\space settings as evidenced by an average increase of $1.12$ in MSE. Finally, in \cref{fig:t2t_ood}, we turn to an NT5-Gen model in a generative setting. As an evaluation metric, we consider \EM\space and observe a decrease of $0.18$ in \EM\space on average.

\section{Application: Overlooked Victim Types}\label{sec:application}
Most event datasets feature only one column detailing victim counts. This column typically quantifies fatalities, as they are considered least ambivalent and most important~\cite{kalyvas_logic_2006, chaudoin_beyond_2017}. The Armed Conflict Location \& Event Data Project (ACLED)~\cite{raleigh_introducing_2010, raleigh_comparing_2019} recently published curated datasets containing violence against healthcare workers, media personnel, and women. Considering the \href{https://acleddata.com/curated-data-files/}{ACLED dataset on Political Violence Targeting Women \& Demonstrations Featuring Women}, we find that more than \num{85}\% of events have \emph{zero} fatalities. This means, many forms of violence remain non-quantified, often those against ``marginalized'' groups of society.

Using the methods presented in this work, we can extract much more fine-grained victim types such as ``injured women'' and ``abducted women''. To this end, we rely on the NT5-Gen model that we fine-tuned on the NAVCO data, without specifically asking for ``women''. In \cref{fig:app}, we present exemplary two-month time series of events in Syria. We find that our model has a higher recall than precision on the ground truth annotations for fatality counts. This may be desirable since we would like to avoid overlooking true victim counts.

\begin{table*}[htb!]
\centering
\resizebox{\textwidth}{!}{
\renewcommand{\arraystretch}{1.3}
\begin{tabular}{lcccccccc}
\hline
  & \multicolumn{3}{c}{Accuracy Optimization} & Reliability & \multicolumn{2}{c}{Robustness} & Hardware\\
   & Absolute Error & Relative Error & String Match & & Need Training & Stable in OOD & \\
\hline
 REGEX & High & Medium & Medium & N/A & No & N/A & Low \\ 
  DEP & Medium & High & Low-Medium & N/A & No & N/A & Low \\ 
  SRL  & Low-Medium & Medium & High & N/A & No & N/A & Low-Medium\\ 
\hline
CLF & N/A & N/A & N/A & Low & Medium-High & Low & Medium - High \\
REG & Low & Low & N/A & Low & High & Low-Medium & Medium - High\\
GEN & Low & Low & Medium-High & High & Low-Medium & Medium-High & High\\
\hline
\end{tabular}}

\caption{Overview of pros and cons of different models. We list baselines: regular expressions~(\texttt{REGEX}), dependency parsing~(\texttt{DEP}), and semantic role labeling~(\texttt{SRL}). The \texttt{CLF}, \texttt{REG}, \texttt{GEN} refer to the fine-tuned NT5-Clf, NT5-Reg, and NT5-Gen models. 
\texttt{Absolute / Relative Error} pertains to the absolute/relative error between true victim counts and model predictions taking the real numerical value of the counts (e.g., mean squared error). \texttt{String Match} considers string metrics like \EM\space used in question answering. The reliability column is based on experiments in model calibration. Robustness is divided into the need for training on a large annotated dataset and the stability in out-of-distribution~(\texttt{OOD}). \texttt{N/A} means ``Not Applicable''.}
\label{tab:summary}
\end{table*}

\section{Discussion}\label{sec:disccusion} 
This work surveys different task formulations of victim count extraction and inspects desiderata like accuracy, reliability, and robustness of different models. We now summarize our findings and conclude which approach performs best under which circumstances~(\cref{tab:summary}). 

Some of the parsing-based approaches have the advantage of requiring no ground truth annotations of the extracted victim counts. This means, there is no need for training, but instead, a manually curated list of patterns and rules has to be assembled. The regex approach, for instance, has minimum requirements regarding hardware, but writing regex patterns is very time-intensive and can be prone to mistakes. Overall, the baseline models shine when it comes to speed, and they perform reasonable when victim counts are explicitly mentioned. Yet they fail at complex reasoning. For instance, when asking for the count of deaths in ``one child and four women lost their lives'', all baselines mistakenly output ``\num{1}''. 

This is where language model-based methods have a competitive edge. The fine-tuned NT5-Gen model has high accuracy both in \EM\space metric and relative error metric. Surprisingly, it is also well-calibrated and relatively robust in the few-shot and out-of-distribution setting. This performance comes at the costs of reduced speed, the requirement of large amounts of training data, and the need for resources like GPUs to be deployed on a large scale. 

Comparing classification and regression objectives, we conclude that classification is easier to handle. In most settings, it may be sufficient to extract a range rather than an exact number anyways. In comparison to generation, in classification and regression settings, models show higher calibration errors and require post-hoc calibration to adjust the model confidence. 

\section{Related Works}\label{sec:related_works}
This work interfaces with related works from different disciplines to improve the measurement of crisis intensity. It draws inspiration from recent advancements in question answering models with a focus on numbers and math word problems. This includes number-enhanced language models more generally. Our work also connects with model calibration in natural language processing (NLP) more generally. 

\paragraph{Measurement of Crisis Intensity.} 
Extracting information about crises has been widely explored using social media data~\cite{Temnikova_Castillo_2015} and newspapers~\cite{keith_identifying_2017,halterman_corpus-level_2021}. 
Most existing measures of crisis intensity focus on counts of event types \cite{goldstein_conflict-cooperation_1992, terechshenko_hot_2020, stoehr_ordinal_2022} or fatality counts \cite{kalyvas_logic_2006}. Previous work studies friend-enemy relationships \cite{han_no_2019, russo_disentangling_2022, stoehr_classifying_2021, stoehr_ordered_2023} and conflict-indicative changes in word embeddings \cite{kutuzov_tracing_2017}.

\paragraph{Numerical Question Answering.}
Numerical Question Answering pertains to the task of providing numeric answers to questions. An exemplary model is NAQANet~\cite{dua_drop_2019}, which extends QANet~\citep{yu_qanet_2018} with numerical operations. Neural Module Networks~\citep{nmn:iclr20} learn and execute a chain of logical learnable and differentiable modules. Some of these modules are specifically targeted at mathematical operations such \texttt{find-num} or \texttt{count}. Other approaches leverage knowledge graphs~\cite{Davidov_Rappoport_2010,kotnis2019learning} or graph neural networks~\cite{chen-etal-2020-question}.
\citet{thawani_representing_2021} provides a detailed overview over methods for representing and modeling numbers in NLP.

\paragraph{Number-enhanced Language Models.}

More recent work in number question answering relies on pre-trained large language models. GenBERT~\cite{geva-etal-2020-injecting} improves numeric reasoning abilities by including a large amount of synthetic data containing numbers. Codex \cite{chen_evaluating_2021} and NT5 \cite{Yang2021NT5TT} apply similar strategies and are trained on code and math word problems. Other approaches focus on step-by-step reasoning such as Minerva \cite{lewkowycz_solving_2022}, scratchpad \cite{nye_show_2022} and chain-of-thought prompting \cite{wei_chain--thought_2022}. \citet{lefebvre_rethinking_2022} propose a prompting-based method particularly for conflict event classification.

\paragraph{Calibration of NLP Models.}

The calibration of NLP models has been extensively studied in classification~\cite{guo_calibration_2017} and structured prediction tasks~\citep{volodymyr2015, nguyen-oconnor-2015-posterior}. Calibration methods have been adapted in language modeling~\citep{pmlr-v119-braverman20a, kong-etal-2020-calibrated}, question answering~\citep{ kamath-etal-2020-selective, Jiang_Araki_Ding_Neubig_2021}, and machine translation~\citep{Kumar2019CalibrationOE, wang-etal-2020-inference}.

\section{Conclusion}
We presented \emph{victim count extraction}, a challenging and impactful task. The task can be tackled using different formulations and models. Models should be evaluated along different dimensions such as accuracy, reliability, and robustness. We survey this ambiguity of victim count extraction, identify promising directions, and discuss outlooks and applications.

\section*{Acknowledgments}

We would like to thank and acknowledge ideas, input, support and feedback from Leonie Muggenthaler, Ryan Cotterell as well as the anonymous reviewers. Niklas Stoehr is supported by a scholarship from the Swiss Data Science Center (SDSC).

\section*{Limitations}
The models may be biased or reproduce biases inherent in their training data. Presenting unrelated, faulty or immoral questions to a model can cause unguided and malicious behavior. For example, we caution of asking questions such as ``How many people \emph{will be injured}...?''; and even worse ``How many people \emph{should be injured}...?''. Improving model calibration will help defending against these issues and enable awareness of when to abstain from answering.

\section*{Ethics Statement}
This work originated from the motivation to diversify victim count extraction towards underrepresented victim types and overlooked forms of violence. This work ultimately intends to assist researchers and analysts in the sector of humanitarian aid who are in demand of accurate victim count information.

\bibliography{anthology}
\bibliographystyle{acl_natbib}

\appendix
\section{Regex Patterns}\label{app:regex}
We convert any non-digitized numeral expressions into a digitized format~(e.g. twelve $\rightarrow$ 12). 
Regex patterns are designed for both passive and active voices.
We also distinguish plural~(``are'' and ``were'') and singular forms~(``is'', ``was'') for passive voice patterns.
The algorithm checks with the following order: passive plural, passive single, and active. 
If multiple numbers are extracted, the first is kept.
We list the regex patterns used to extract victim counts in \cref{tab:regex_patterns}, for death counts and injury counts respectively.

\begin{table*}[hbt!]
\fontsize{10}{10}\selectfont
\centering
\renewcommand{\arraystretch}{1.55} 
\setlength{\tabcolsep}{0.35em} 
\resizebox{\textwidth}{!}{%
\begin{tabular}{@{}lll@{}}
\hline
 Data Type & Regex Type &  Regex Pattern \\
\hline
Death  & Passive Plural &
  \texttt{\textbackslash d(\textbackslash d|,)*(?!\textbackslash D*(injur|wound))(?=.*(\textbackslash b(were|are)\textbackslash D*\textbackslash b(killed|dead|died|slain)))
  }\\ 
 & Passive Singular & \texttt{\textbackslash S*(?!\textbackslash D*(injur|wound))(?=.*(\textbackslash b(was|is)\textbackslash D*\textbackslash b(killed|dead|died|slain)))} \\ 
 & Active & \texttt{(kill|slay|slain)\textbackslash D*\textbackslash b\textbackslash d(\textbackslash d|,)*} \\
  \hline
Injury  & Passive Plural & \texttt{\textbackslash d(\textbackslash d|,)*(?!.*(\textbackslash b(were|are)?\textbackslash D*\textbackslash b(killed|dead|died|slain)))(?=.*\textbackslash b(injur|wound))} \\ 
 & Passive Singular & \texttt{\textbackslash S*(?=(was|is).*\textbackslash b(injur|wound))(?!\textbackslash D*(\textbackslash b(were|are)\textbackslash D*\textbackslash b(killed|dead|died|slain)))} \\
 & Active & \texttt{(injured?|wound)\textbackslash D*\textbackslash d+} \\
  \hline
\end{tabular}%
}

\caption{Regex patterns.}
\label{tab:regex_patterns}
\end{table*}

\section{Accuracy Evaluation}\label{app:accuracy}
In this section, we complement the accuracy evaluation of the models in \cref{sec:accuracy}. 

\subsection{\EM\space and \fone\space score on Death Counts}
The \EM\space and \fone\space scores on extracting the death counts are shown in \cref{tab:em_f1_death}, which compares the performance of the baseline models and the fine-tuned NT5-Gen model. Similar to the results on the injury counts~\cref{tab:em_f1_injury}, the fine-tuned NT5-Gen model performs better than all baselines and the SRL has the best accuracy among baselines. 

\begin{table}[htb]
\fontsize{10}{10}\selectfont
\centering
\renewcommand{\arraystretch}{1.55} 
\setlength{\tabcolsep}{0.35em} 
\resizebox{\columnwidth}{!}{
\renewcommand{\arraystretch}{1.3}
\begin{tabular}{lrrrrrr}
\hline
 \multicolumn{1}{l}{} & \multicolumn{3}{c}{Exact Match} & \multicolumn{3}{c}{\fone} \\
\cline{2-7}
 \multicolumn{1}{l}{} & WAD & NAVCO & EMM & WAD & NAVCO & EMM \\ 
\hline
 Regex & 0.3543 & 0.3921 & 0.2835 & 0.3897 & 0.4196 & 0.3242 \\ 
  Dep & 0.1506 & 0.3526 & 0.0767 & 0.2064 & 0.3792 & 0.1317 \\ 
  SRL & 0.4342 & 0.4839 & 0.3972 & 0.7794 & 0.4837 & 0.3613\\ 
\hline
NT5 & \textbf{0.6798} & \textbf{0.6590} & \textbf{0.6322} & \textbf{0.8458} & \textbf{0.5436} & \textbf{0.4917} \\
\hline
\end{tabular}}
\caption{\EM\space and \fone\space scores of the baseline models and the fine-tuned NT5-Gen model on death counts. Best metrics are \textbf{bolded}. \texttt{DEP} refers to the dependency parsing model and \texttt{SRL} refers to the semantic role labeling model.}
\label{tab:em_f1_death}
\end{table}

\subsection{Confusion Matrix on Death Counts}
\begin{figure*}[!hbt]
    \centering
    \resizebox{\textwidth}{!}{
    \includegraphics{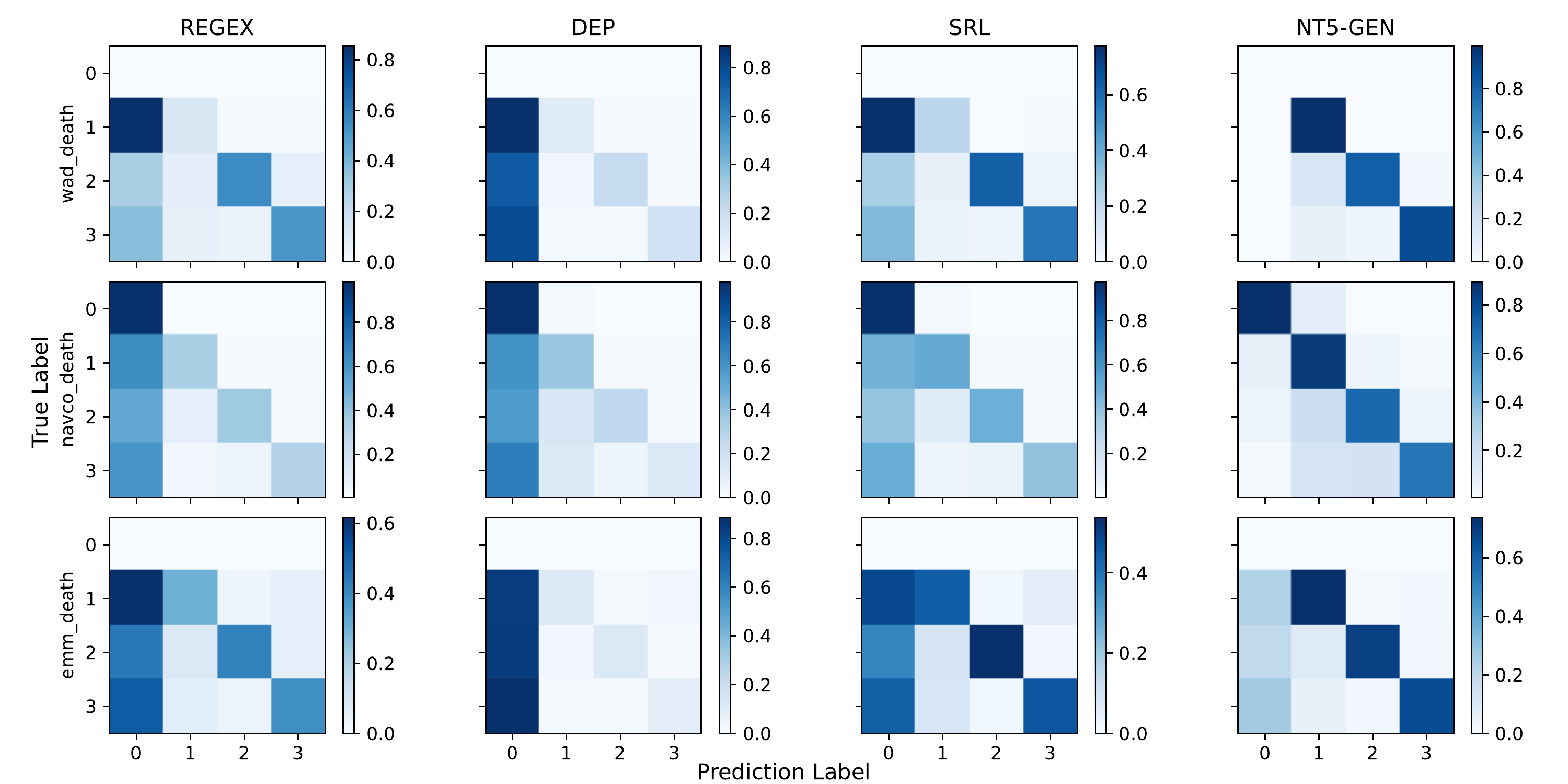}}
    \caption{Confusion matrices of baseline models and fine-tuned NT5-Gen model (columns) of extracting death counts from different data~(rows). We convert the true \vc and model predictions into 4 categories: for any count $y$, ``0'' is $y = 0$, ``1'' is $0< y \leq 3$, ``2'' is $3 < y \leq 10$ and ``3'' is $y > 10$. Values are normalized over true counts. 
}
    \label{fig:confusion_matrix_death}
\end{figure*}

Similar to \cref{fig:confusion_matrix_injury} shown in \cref{subsec:compare_baseline}, \cref{fig:confusion_matrix_death} plots the confusion matrices of the binned death counts for the different datasets, which compare the accuracy of the baseline models with the fine-tuned NT5-Gen model.

\subsection{Results on Classification and Regression} 
In \cref{subsec:results-clf-regression}, we have shown the results of the NT5-Clf model and the NT5-Reg model fine-tuned on NAVCO injury counts in \cref{tab:classification_accuracy}. 

Here, we use the same metrics and display the classification performance on other datasets. In specific, \cref{tab:wad_death_clf}, \cref{tab:wad_injury_clf}, \cref{tab:navco_death_clf}, \cref{tab:emm_death_clf}, and \cref{tab:emm_injury_clf} respectively show the classification performance of the NT5-Clf model fine-tuned on WAD death counts, WAD injury counts, NAVCO death counts, EMM death counts, and EMM injury counts.

\begin{figure}[htb!]
    \centering
    \resizebox{\columnwidth}{!}{
    \includegraphics{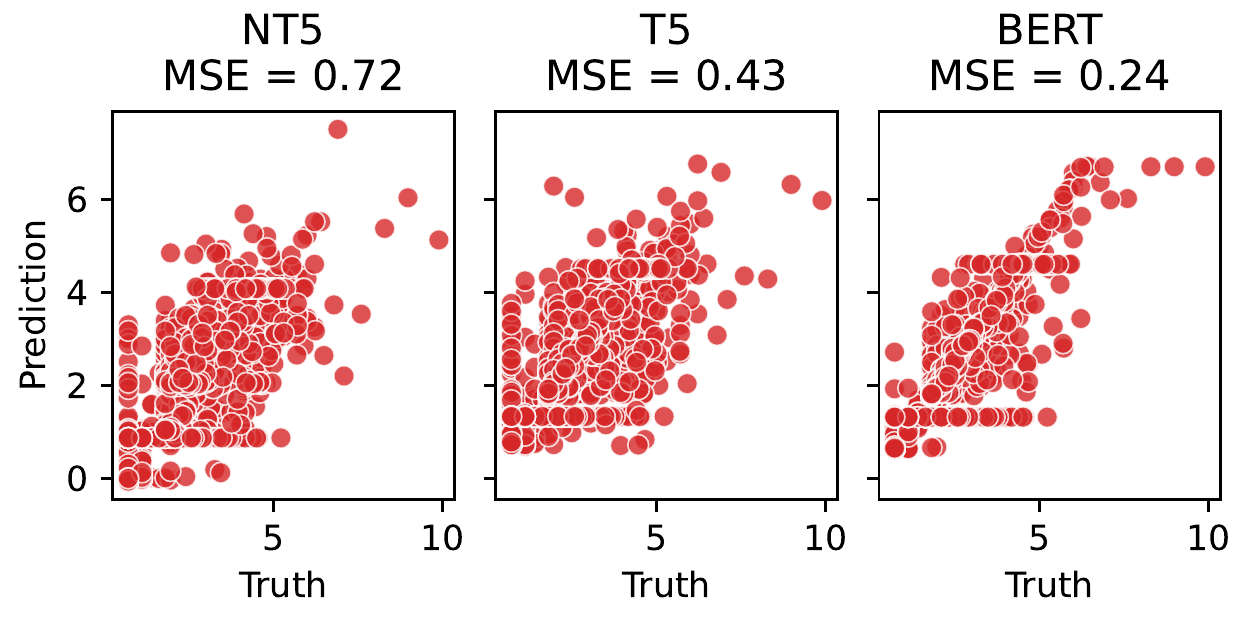}}
    \caption{Scatter plots of the fine-tuned NT5-Reg model initialized with different pre-trained weights (\textsc{nt5}, \textsc{t5-small}, and \textsc{bert-base-uncased}). The models are trained on log-transformed victim counts of WAD death.}
    \label{fig:regression_wad_death}
\end{figure}

\begin{figure}[htb!]
    \centering
    \resizebox{\columnwidth}{!}{
    \includegraphics{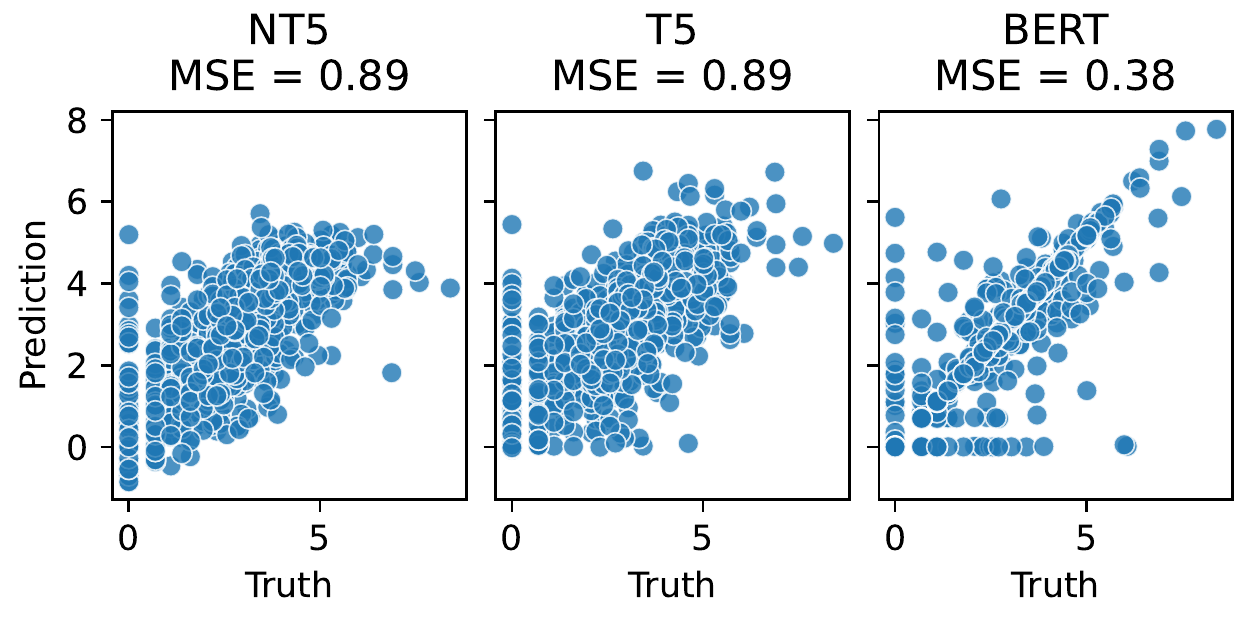}}
    \caption{Scatter plots of the fine-tuned NT5-Reg model initialized with different pre-trained weights (\textsc{nt5}, \textsc{t5-small}, and \textsc{bert-base-uncased}). The models are trained on log-transformed victim counts of WAD injury.}
    \label{fig:regression_wad_injury}
\end{figure}

\begin{figure}[htb!]
    \centering
    \resizebox{\columnwidth}{!}{
    \includegraphics{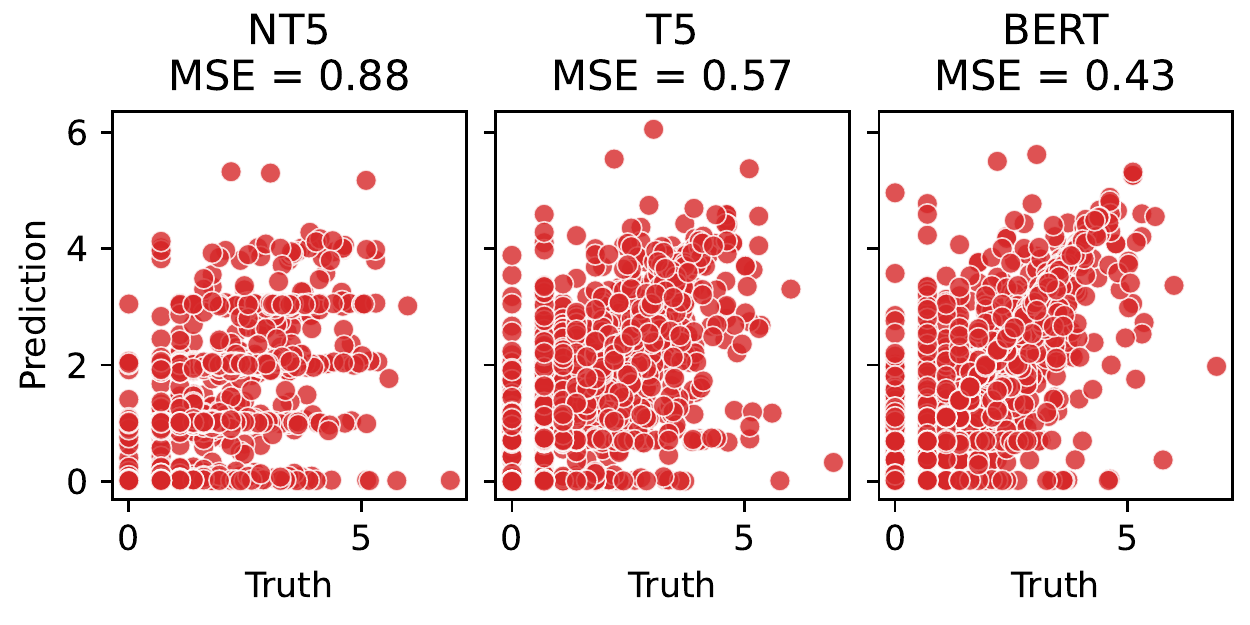}}
    \caption{Scatter plots of the fine-tuned NT5-Reg model initialized with different pre-trained weights (\textsc{nt5}, \textsc{t5-small}, and \textsc{bert-base-uncased}). The models are trained on log-transformed victim counts of NAVCO death.}
    \label{fig:regression_navco_death}
\end{figure}

\begin{figure}[htb!]
    \centering
    \resizebox{\columnwidth}{!}{
    \includegraphics{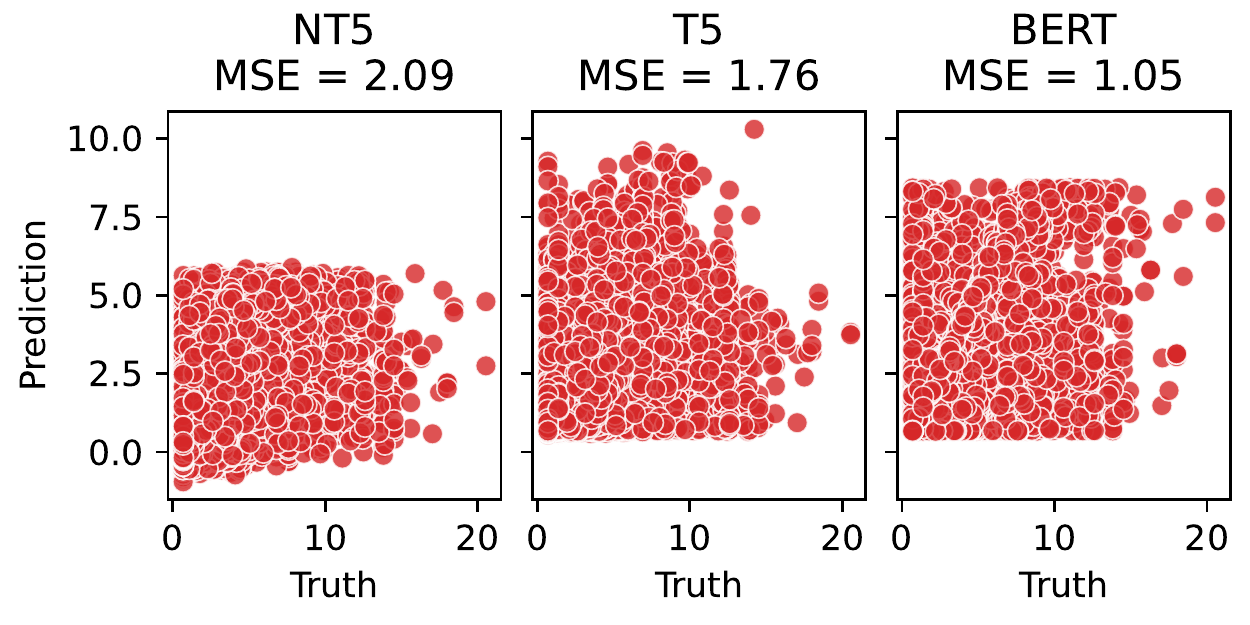}}
    \caption{Scatter plots of the fine-tuned NT5-Reg model initialized with different pre-trained weights (\textsc{nt5}, \textsc{t5-small}, and \textsc{bert-base-uncased}). The models are trained on log-transformed victim counts.}
    \label{fig:regression_emm_death}
\end{figure}

\begin{figure}[htb!]
    \centering
    \resizebox{\columnwidth}{!}{
    \includegraphics{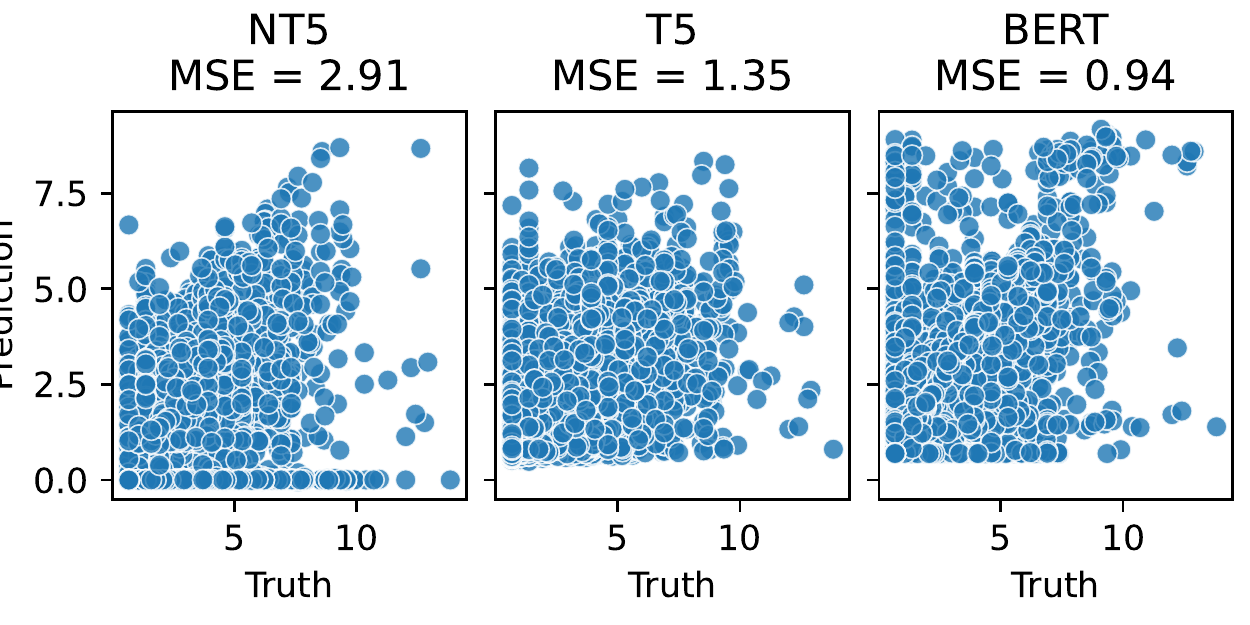}}
    \caption{Scatter plots of the fine-tuned NT5-Reg model initialized with different pre-trained weights (\textsc{nt5}, \textsc{t5-small}, and \textsc{bert-base-uncased}). The models are trained on log-transformed victim counts.}
    \label{fig:regression_emm_injury}
\end{figure}

Similarly, we provide the scatter plots of the fine-tuned NT5-Reg models initialized with different pre-trained weights in this section: WAD death counts~(\cref{fig:regression_wad_death}), WAD injury counts~(\cref{fig:regression_wad_injury}), NAVCO death counts~(\cref{fig:regression_navco_death}), EMM death counts~(\cref{fig:regression_emm_death}), and EMM injury counts~(\cref{fig:regression_emm_injury}).

\begin{table}[htb!]
\fontsize{11}{11}\selectfont
\centering
\renewcommand{\arraystretch}{1.} 
\setlength{\tabcolsep}{0.35em} 
\resizebox{\columnwidth}{!}{
\begin{tabular}{lrrrr}
\toprule
  & Accuracy & F1 score & Precision & Recall\\
  \midrule
  \texttt{NT5} & 0.81 & 0.81 & 0.80 & 0.83\\
  \texttt{T5} & 0.81 & 0.81 & 0.81 & 0.84 \\
  \texttt{BERT} & 0.86 & 0.86 & 0.86 & 0.88\\
\bottomrule
\end{tabular}}
\caption{Classification results on WAD death counts with the NT5-Clf model initialized by different pre-trained weights: \textsc{nt5}, \textsc{t5-small}, and \textsc{bert-base-uncased}. \fone\space, precision and recall scores are macro.}
\label{tab:wad_death_clf}
\end{table}

\begin{table}[htb!]
\fontsize{11}{11}\selectfont
\centering
\renewcommand{\arraystretch}{1.} 
\setlength{\tabcolsep}{0.35em} 
\resizebox{\columnwidth}{!}{
\begin{tabular}{lrrrr}
\toprule
  & Accuracy & F1 score & Precision & Recall\\
  \midrule
  \texttt{NT5} & 0.77 & 0.69 & 0.70 & 0.69 \\
  \texttt{T5} & 0.76 & 0.69 & 0.70 & 0.68 \\
  \texttt{BERT} & 0.93 & 0.91 & 0.91 & 0.90\\
\bottomrule
\end{tabular}}
\caption{Classification results on WAD injury counts with the NT5-Clf model initialized by different pre-trained weights: \textsc{nt5}, \textsc{t5-small}, and \textsc{bert-base-uncased}. \fone\space, precision and recall scores are macro.}
\label{tab:wad_injury_clf}
\end{table}

\begin{table}[htb!]
\fontsize{11}{11}\selectfont
\centering
\renewcommand{\arraystretch}{1.} 
\setlength{\tabcolsep}{0.35em} 
\resizebox{\columnwidth}{!}{
\begin{tabular}{lrrrr}
\toprule
  & Accuracy & F1 score & Precision & Recall\\
  \midrule
  \texttt{NT5} & 0.65 & 0.60 & 0.62 & 0.59\\
  \texttt{T5} & 0.65 & 0.60 & 0.61 & 0.59\\
  \texttt{BERT} & 0.52 & 0.23 & 0.17 & 0.33\\
\bottomrule
\end{tabular}}
\caption{Classification results on NAVCO death counts with the NT5-Clf model initialized by different pre-trained weights: \textsc{nt5}, \textsc{t5-small}, and \textsc{bert-base-uncased}. \fone\space, precision and recall scores are macro.}
\label{tab:navco_death_clf}
\end{table}

\begin{table}[htb!]
\fontsize{11}{11}\selectfont
\centering
\renewcommand{\arraystretch}{1.} 
\setlength{\tabcolsep}{0.35em} 
\resizebox{\columnwidth}{!}{
\begin{tabular}{lrrrr}
\toprule
  & Accuracy & F1 score & Precision & Recall\\
  \midrule
  \texttt{NT5} & 0.72 & 0.65 & 0.66 & 0.65\\
  \texttt{T5} & 0.70 & 0.63 & 0.65 & 0.63\\
  \texttt{BERT} & 0.84 & 0.80 & 0.82 & 0.78\\
\bottomrule
\end{tabular}}
\caption{Classification results on EMM death counts with the NT5-Clf model initialized by different pre-trained weights: \textsc{nt5}, \textsc{t5-small}, and \textsc{bert-base-uncased}. \fone\space, precision and recall scores are macro.}
\label{tab:emm_death_clf}
\end{table}

\begin{table}[htb!]
\fontsize{11}{11}\selectfont
\centering
\renewcommand{\arraystretch}{1.} 
\setlength{\tabcolsep}{0.35em} 
\resizebox{\columnwidth}{!}{
\begin{tabular}{lrrrr}
\toprule
  & Accuracy & F1 score & Precision & Recall\\
  \midrule
  \texttt{NT5} & 0.68 & 0.58 & 0.60 & 0.57\\
  \texttt{T5} & 0.68 & 0.58 & 0.59 & 0.57\\
  \texttt{BERT} & 0.81 & 0.77 & 0.79 & 0.76\\
\bottomrule
\end{tabular}}
\caption{Classification results on EMM injury counts with the NT5-Clf model initialized by different pre-trained weights: \textsc{nt5}, \textsc{t5-small}, and \textsc{bert-base-uncased}. \fone\space, precision and recall scores are macro.}
\label{tab:emm_injury_clf}
\end{table}


\section{Robustness Evaluation}
In this section, we provide the detailed performance of the few-shot setting~(\cref{subsec:app_fewshot}) and the out-of-distribution setting~(\cref{subsec:app_ood}) discussed in \cref{sec:robustness}. 

\subsection{Few-shot Performance}\label{subsec:app_fewshot}
\begin{figure}[htb!]
    \centering
    \resizebox{\columnwidth}{!}{
    \includegraphics{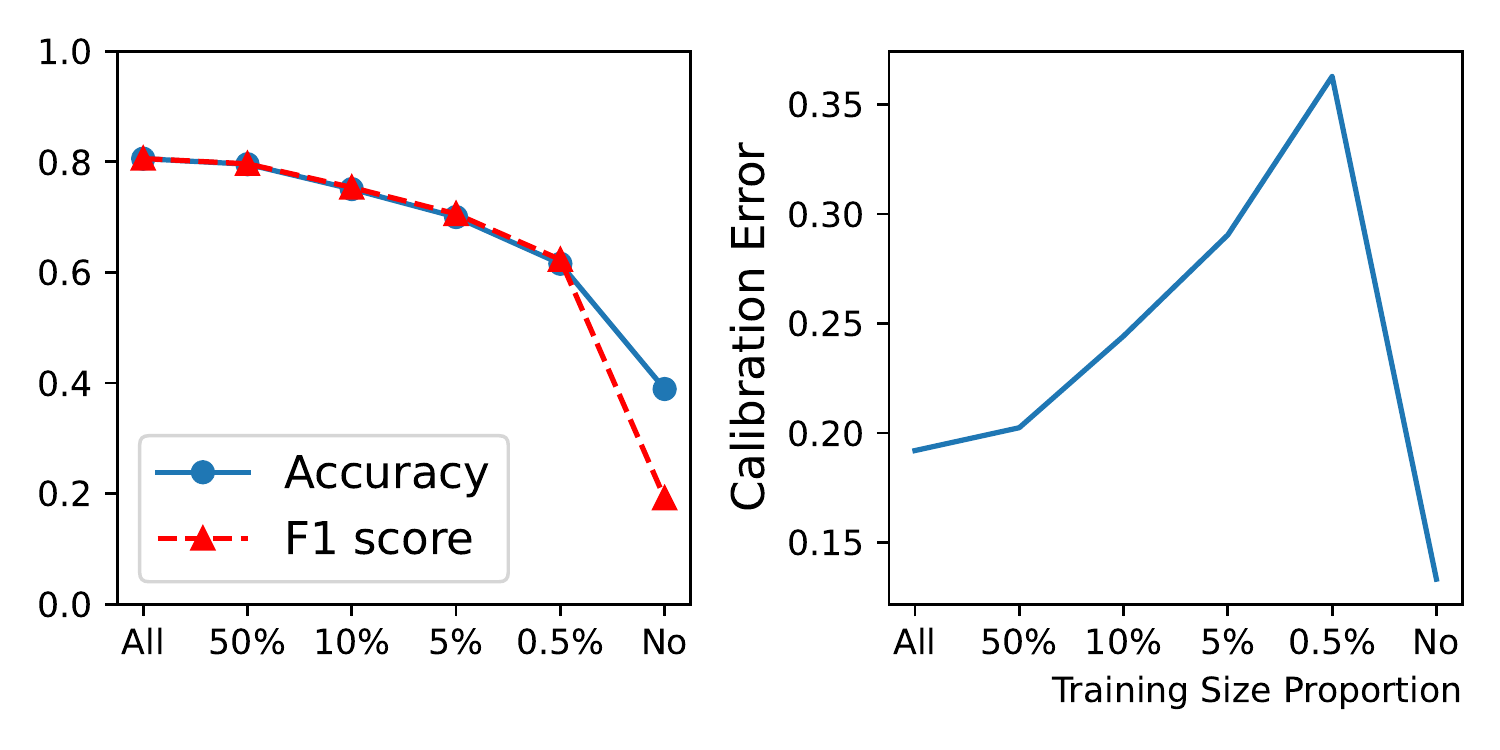}
    }
    \caption{Few-shot performance of the fine-tuned classification model on WAD death counts.}
    \label{fig:fewshot_clf}
\end{figure}

\begin{figure}[htb!]
    \centering
    \resizebox{\columnwidth}{!}{
    \includegraphics{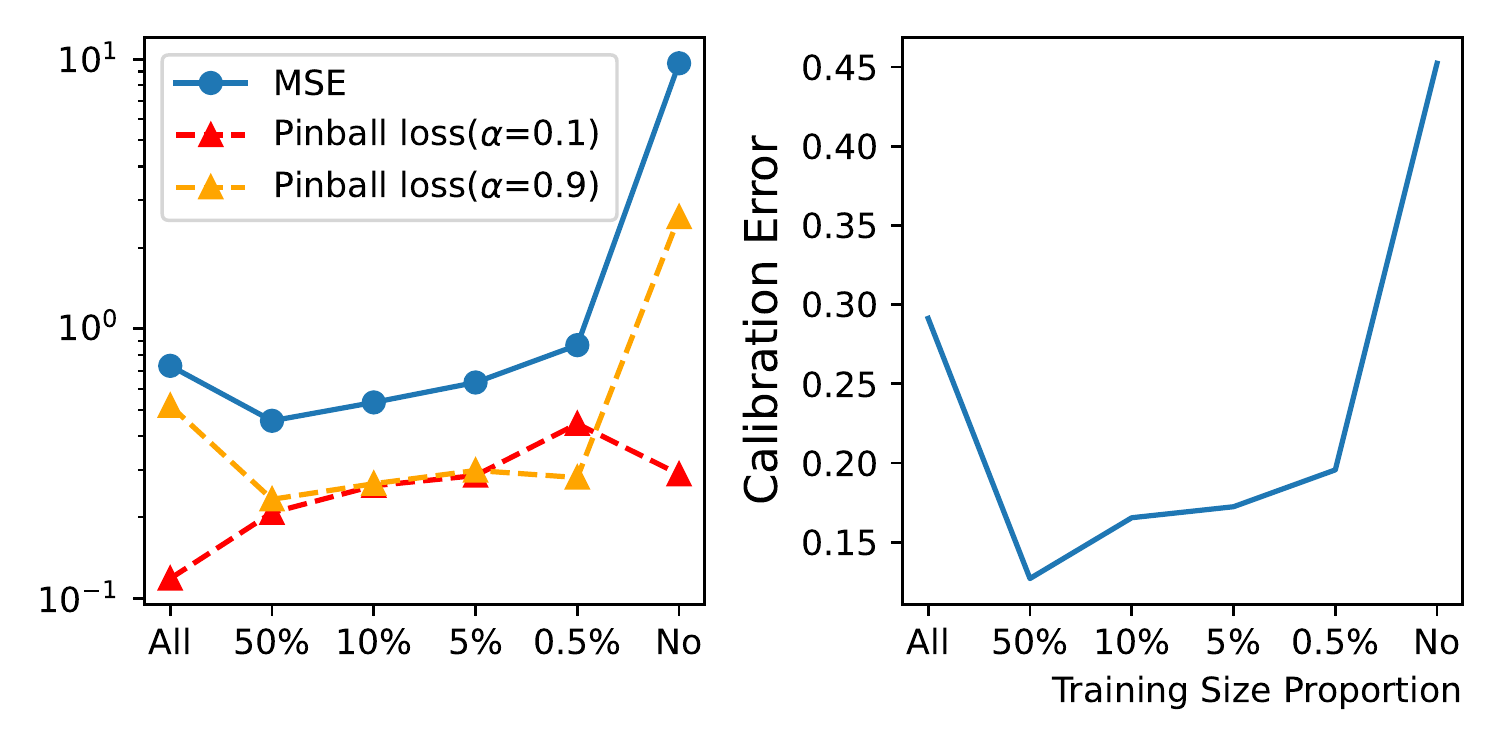}
    }
    \caption{Few-shot performance of the fine-tuned regression model on WAD death counts.}
    \label{fig:fewshot_reg}
\end{figure}

\begin{figure}[ht]
    \centering
    \resizebox{\columnwidth}{!}{
    \includegraphics{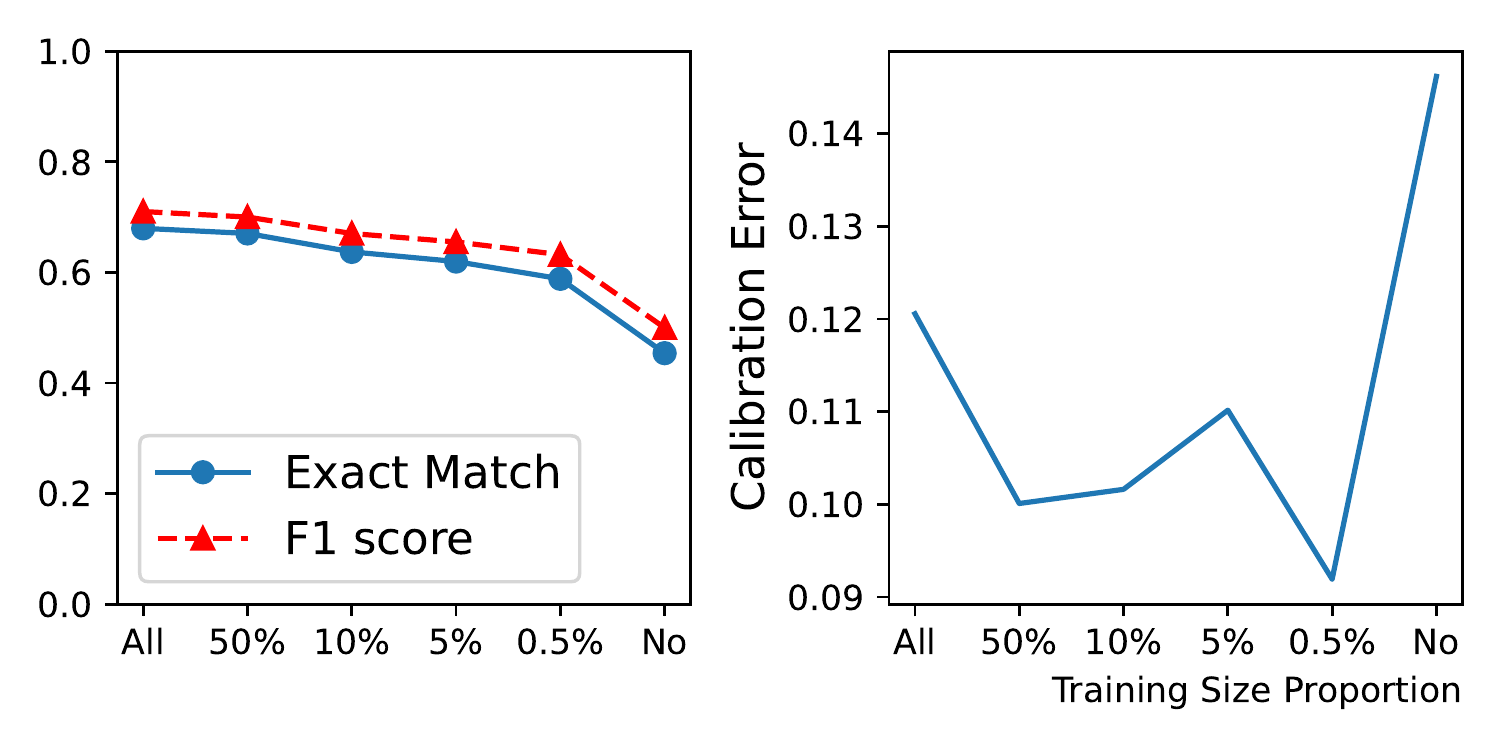}
    }
    \caption{Few-shot performance of the fine-tuned NT5-Gen model on WAD death counts.}
    \label{fig:fewshot_t2t}
\end{figure}

We display the results of the few-shot settings where different proportions of the training set are used to fine-tune the models.
For each formulation, the left figure is the variation of the accuracy metrics and the right figure is the variation of the calibration error. 
\cref{fig:fewshot_clf}, \cref{fig:fewshot_reg}, and \cref{fig:fewshot_t2t} are performance of the few-shot settings of the fine-tuned NT5-Clf, NT5-Reg, and NT5-Gen models respectively. 

With respect to accuracy metrics, the classification accuracy and the \fone\space score is plotted for the fine-tuned NT5-Clf model. 

For the regression, we plot the change of mean squared error on the $\log$ transformed counts. In addition, we plot the pinball losses using two targeting quantile (at 10\% and at 90\%).

Lastly, the \EM\space and the \fone\space scores are drawn for the fine-tuned NT5-Gen model.

\subsection{Out-of-distribution Setting}\label{subsec:app_ood}
\begin{figure}[htb!]
    \centering
    \resizebox{\columnwidth}{!}{
    \includegraphics{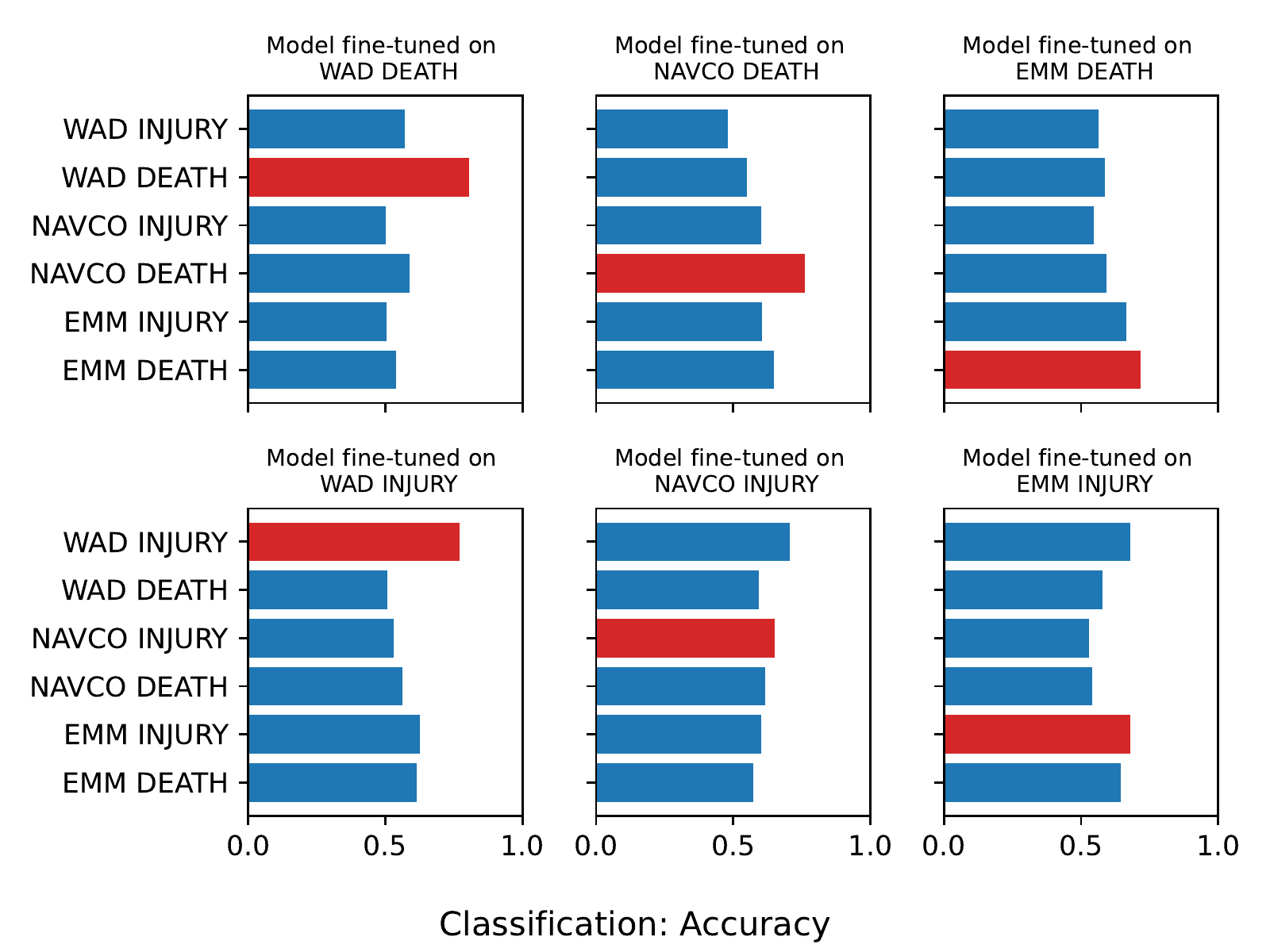}}
    \caption{Classification accuracy for using the fine-tuned NT5-Clf models on out-of-distribution data~(blue) and on in-distribution data~(red)}
    \label{fig:clf_ood}
\end{figure}

\begin{figure}[htb!]
    \centering
    \resizebox{\columnwidth}{!}{
    \includegraphics{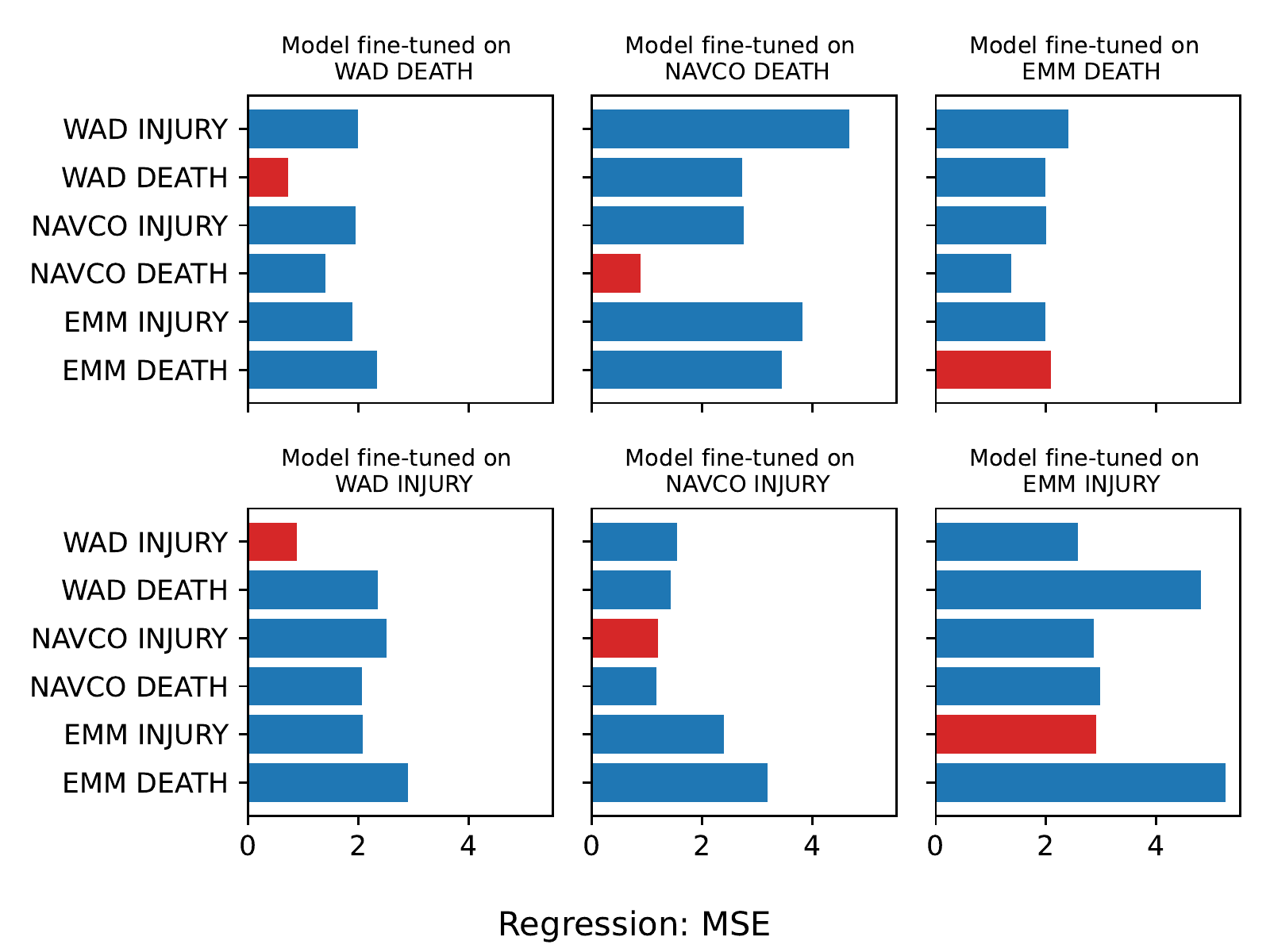}}
    \caption{Mean squared error for using the fine-tuned NT5-Reg models on out-of-distribution data~(blue) and on in-distribution data~(red)}
    \label{fig:reg_ood}
\end{figure}

\begin{figure}[htb!]
    \centering
    \resizebox{\columnwidth}{!}{
    \includegraphics{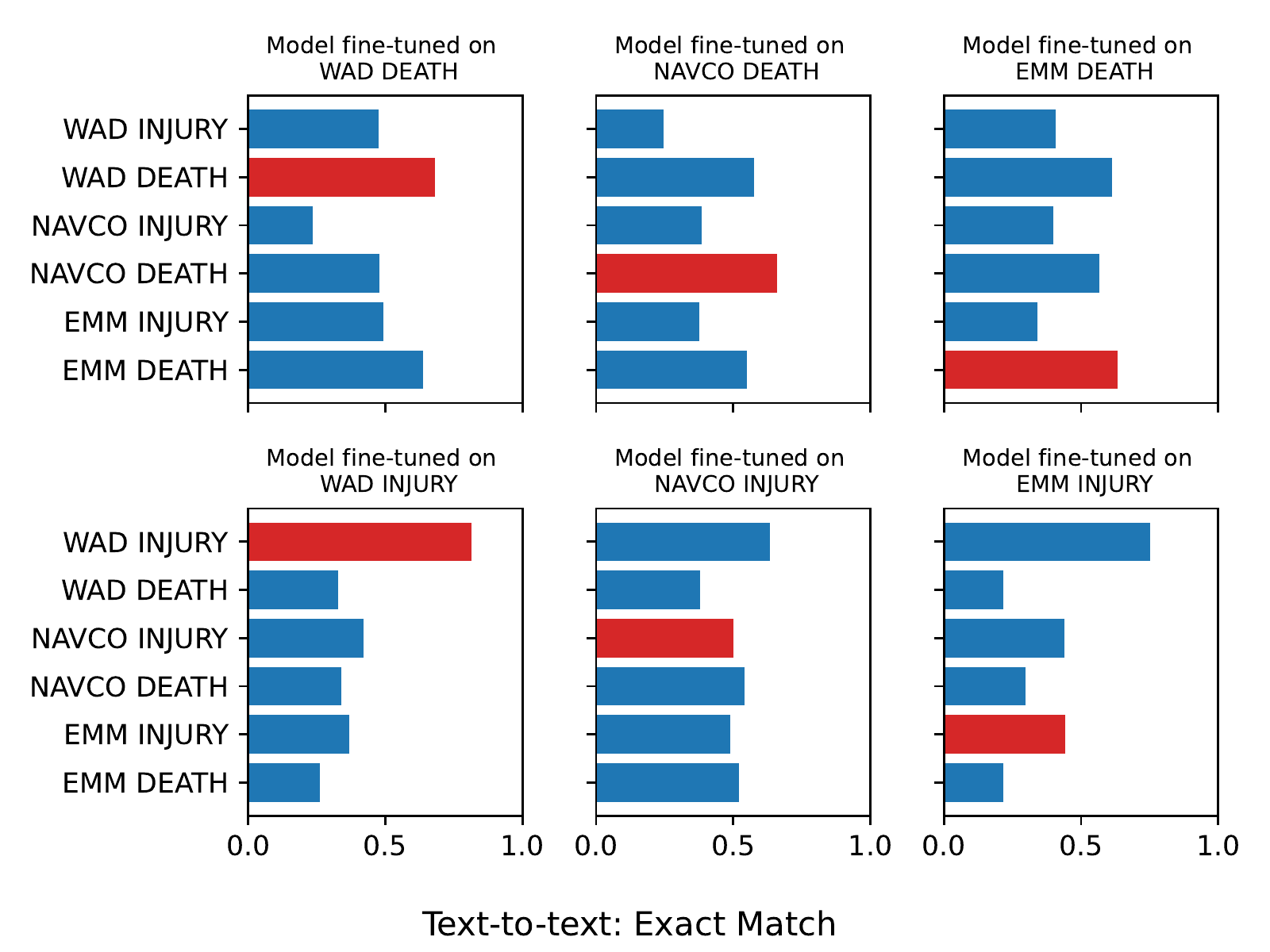}}
    \caption{\EM\space for using the fine-tuned NT5-Gen models on out-of-distribution data~(blue) and on in-distribution data~(red)}
    \label{fig:t2t_ood}
\end{figure}

For each task formulation, we examine the accuracy performance in the \ood setting for the fine-tuned NT5-Clf~(\cref{fig:clf_ood}), NT5-Reg~(\cref{fig:reg_ood}), and NT5-Gen~(\cref{fig:t2t_ood}).
For all plots, the x-axis is the accuracy metric used in each task formulation, and the y-axis indicates the test set to be made inferences on. 
The red bar indicates the performance of in-distribution performance, e.g., accuracy of WAD death test data using the model fine-tuned on WAD death.

With respect to the accuracy metric, different formulations use their corresponding metric. 
For the classification setting, we show the variation in classification accuracy. 
For the regression setting, we show the variation in mean squared errors.
For the generation setting, we show the change in \EM\space scores. 

\end{document}